\documentclass{article}

\PassOptionsToPackage{numbers, compress}{natbib}

\usepackage[final]{neurips_2025}

\usepackage[utf8]{inputenc} %
\usepackage[T1]{fontenc}    %
\usepackage{hyperref}       %
\usepackage{url}            %
\usepackage{booktabs}       %
\usepackage{amsfonts}       %
\usepackage{nicefrac}       %
\usepackage{microtype}      %
\usepackage{xcolor}         %

\usepackage{abbrv}

\usepackage{amsmath}
\usepackage{multirow}
\usepackage{wrapfig}
\usepackage{graphicx}
\usepackage{tabularx}
\usepackage{caption}
\usepackage{subcaption}
\usepackage[export]{adjustbox}
\usepackage[capitalize]{cleveref}
\crefname{figure}{Figure}{Figures}

\usepackage[most]{tcolorbox}
\definecolor{prompt}{HTML}{5f84e4}
\definecolor{img}{HTML}{820100}

\usepackage{listings}
\def\method{\textsc{SpatialLM}\xspace}

\renewcommand{\P}{\mathbf{P}}

\renewcommand{\L}{\mathbf{L}}

\newcommand{\F}{\mathbf{F}}

\renewcommand{\O}{\mathbf{O}}

\newcolumntype{Y}{>{\centering\arraybackslash}X}

\definecolor{seatings}{HTML}{39e133}
\definecolor{beddings}{HTML}{9d2bf5}
\definecolor{cabinetry}{HTML}{fd8630}
\definecolor{tables}{HTML}{33ebf7}
\definecolor{kitchen_appliances}{HTML}{3e6271}
\definecolor{bathroom_fixtures}{HTML}{f786e1}
\definecolor{lighting}{HTML}{c9f659}
\definecolor{decoration}{HTML}{e92c7e}
\definecolor{electronics}{HTML}{5485fe}
\definecolor{others}{HTML}{a4cbdb}

\title{\method: Training Large Language Models for Structured Indoor Modeling}

\author{
Yongsen Mao\textsuperscript{1}\thanks{Equal contribution.} \quad
Junhao Zhong\textsuperscript{1}$^*$ \quad
Chuan Fang\textsuperscript{2} \quad
Jia Zheng\textsuperscript{1} \\
\textbf{Rui Tang}\textsuperscript{1} \quad
\textbf{Hao Zhu}\textsuperscript{1} \quad
\textbf{Ping Tan}\textsuperscript{2} \quad
\textbf{Zihan Zhou}\textsuperscript{1} \\
\textsuperscript{1}Manycore Tech Inc. \quad
\textsuperscript{2}Hong Kong University of Science and Technology \\
\texttt{\url{https://manycore-research.github.io/SpatialLM}} \\
}

\begin{document}

\maketitle

\begin{abstract}

\method is a large language model designed to process 3D point cloud data and generate structured 3D scene understanding outputs. These outputs include architectural elements like walls, doors, windows, and oriented object boxes with their semantic categories. Unlike previous methods which exploit task-specific network designs, our model adheres to the standard multimodal LLM architecture and is fine-tuned directly from open-source LLMs.

To train \method, we collect a large-scale, high-quality synthetic dataset consisting of the point clouds of 12,328 indoor scenes (54,778 rooms) with ground-truth 3D annotations, and conduct a careful study on various modeling and training decisions. On public benchmarks, our model gives state-of-the-art performance in layout estimation and competitive results in 3D object detection. With that, we show a feasible path for enhancing the spatial understanding capabilities of modern LLMs for applications in augmented reality, embodied robotics, and more.

\end{abstract}

\section{Introduction}
\label{sec:intro}

3D indoor environments are ubiquitous in our daily lives. We spend a lot of time and perform various activities in such environments every day. Therefore, a long-standing goal of artificial intelligence is to teach machines to perceive, reason about, and interact with 3D indoor scenes as humans do. In this work, we focus on the task of \emph{structured indoor modeling}, which aims to extract structured indoor scene descriptions from raw sensorial inputs (\ie, RGBD scans). Specifically, the scene description includes both the architectural layout (\ie, walls, doors, and windows) and 3D object bounding boxes in the indoor environment. Such 3D structure information has been shown to benefit a wide range of real-world applications such as scene editing~\cite{PatilPLFSZ24}, augmented reality~\cite{ARKitScenes, SceneScript}, and robot navigation~\cite{GuSWTW22}.

\begin{figure}[t]
    \centering
    \includegraphics[width=\linewidth]{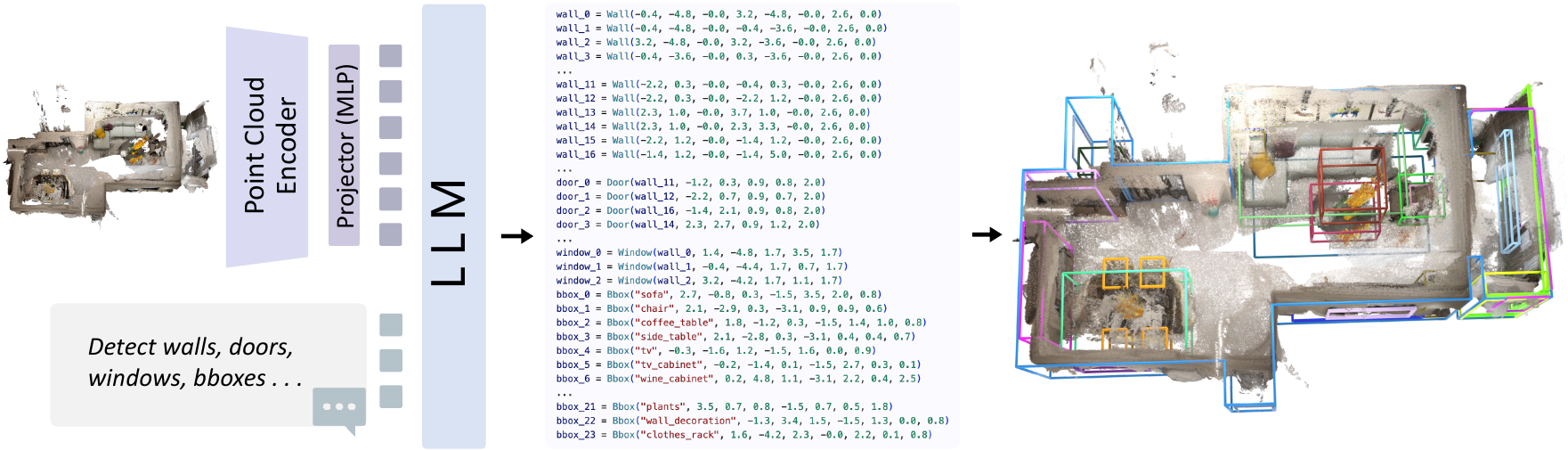}
    \caption{The overall pipeline of \method. Given the point cloud input, it employs a standard ``Encoder-MLP-LLM'' architecture for multimodal feature alignment {\bf (left)}, and generates structured scene descriptions in pure text form as output {\bf (middle)}. The reconstructed 3D structure is further overlaid on the point cloud for visualization {\bf (right)}.}
    \label{fig:pipeline}
\end{figure}

Compared to 3D formats such as meshes, voxels, or implicit functions, the structured scene description provides a highly compact yet flexible scene representation. In this paper, our \emph{first contribution} is to regard the structured descriptions as scripts of a general-purpose language (\ie, Python) and propose to predict the language in text form, as illustrated in \cref{fig:pipeline}. This design choice has several advantages: (i) it is human interpretable and editable; (ii) it can be easily extended to incorporate any new classes without affecting existing content in the scripts; and (iii) it allows us to leverage the strong built-in coding capability of pre-trained large language models. 

Therefore, our goal is to implement our method by directly fine-tuning open-source LLMs. While modern (multimodal) LLMs have revolutionized fields like natural language processing, 2D image understanding and generation, few work has attempted to use LLMs for 3D structured scene modeling. This is partly due to the lack of large-scale, high-quality datasets for the task. In this paper, as our \emph{second contribution}, we collect a new synthetic dataset, which consists of the point clouds of 12,328 distinct scenes (54,778 rooms) paired with 3D structure information, and conduct the first empirical study on the best strategy of aligning the point cloud input with LLMs for structured scene modeling.

Building upon our findings, we introduce \method, a large language model which is capable of processing point cloud input and generating structured scene descriptions. As our \emph{third contribution}, we show that, by first training on our large-scale dataset and then on smaller downstream task data, our model gives competitive performance on public benchmarks for layout estimation and 3D object detection, respectively. Additionally, we provide proof-of-concept zero-shot experiments which show that our model can handle point clouds from diverse sources such as monocular video sequences. These results suggest that \method may serve as a base for developing future solutions for scenarios where enhanced spatial understanding and reasoning is required (\eg, embodied AI).

\section{Related Work}

\subsection{Structured Indoor Modeling}

Structured indoor modeling aims to recover structural elements (\ie, walls, doors, and windows) as well as 3D objects from unstructured point clouds. In the literature, some studies focus on room layout estimation from point clouds. Early methods~\cite{MuraliSOP17,abs-1907-00631} rely on traditional geometric analysis algorithms (\ie, RANSAC) to detect wall planes. SceneCAD~\cite{AvetisyanKCDDN20} jointly optimizes the layout estimation and CAD model alignment of the scanned scene. RoomFormer~\cite{YueKSE23} proposes to train a Transformer-based network to predict corners and rooms. Meanwhile, a number of recent methods~\cite{MisraGJ21, RukhovichV022, WangDDSLLL022, 0001GYLLW00G24, KolodiazhnyiVSR25} have advanced the state-of-the-art on 3D object detection from point clouds. 

The closest work to ours is SceneScript~\cite{SceneScript}, which also casts structured indoor reconstruction as a sequence modeling problem. It demonstrates great flexibility as the language can be tailored to different scenarios without changing the method. But unlike our method, SceneScript uses domain-specific tokens for language commands, thus requiring a specialized decoder. Instead, \method strictly adheres to the ``Encoder-MLP-LLM'' architecture and is directly fine-tuned based on modern foundation models.

\subsection{Large Language Models for 3D Scene Understanding}

Recently, there has been a growing interest in empowering LLMs to understand and reason about 3D spaces, as defined in public benchmarks such as ScanRefer~\cite{ChenCN20} and ScanQA~\cite{AzumaMKK22}. One line of work~\cite{Chat-3D, HuangYMLLW0ZJ024, Chat-Scene, Uni3D-LLM, SceneVerse} first uses 3D detectors to locate objects in the point cloud, then extracts features of individual objects using a pre-trained object-level encoder. In this paper, we study how such 3D detectors can be learned with modern LLMs.

Other methods directly extract scene-level features from the point cloud and align the features with LLMs~\cite{HongZCZDCG23, ZhuWZCL24, ChenCZLY0Z0024, Scene-LLM}. 3D-LLM~\cite{HongZCZDCG23} uses pre-trained image encoder to obtain 2D features from multi-view images and back-projects onto the point cloud~\cite{JatavallabhulaK23}. It then leverages a Q-former~\cite{BLIP2} to map them to a fixed number of tokens as input to the LLM. LL3DA~\cite{ChenCZLY0Z0024} adopts a 3D visual encoder pre-trained on ScanNet detection as the scene encoder. ScanReason~\cite{ZhuWZCL24} further employs a 3D point cloud encoder to obtain fine-grained geometric information for accurate object grounding. Scene-LLM~\cite{Scene-LLM} tokenizes the 3D point cloud features by simply dividing the space into a fixed-resolution voxel grid and aggregating features in the same voxel.

Besides point cloud-based representations, recent studies have explored other ways to integrate 3D information in LLMs. SpatialVLM~\cite{0003XKISGX24} primarily focuses on synthesizing large-scale 3D VQA data to enhance the spatial reasoning capabilities of the VLM. LLaVA-3D~\cite{LLaVA-3D} injects 3D position embeddings into 2D CLIP features to establish a unified architecture for 2D image understanding and 3D scene understanding. SpatialRGPT~\cite{ChengYFGYK0L24} extracts a 3D scene graph from a single image to enhance grounded spatial reasoning in LLMs. Video-3D LLM~\cite{Video-3D-LLM} learns representations from video frames and 3D coordinates to leverage the strong priors of pre-trained 2D Video LLMs. LSceneLLM~\cite{LSceneLLM} uses an adaptive self-attention module to capture fine-grained details in local areas. Our work differs from these studies in that we focus on analyzing the structural information of the scene.

\section{\method}
\label{sec:method}

\begin{figure}[t]
    \centering
    \begin{minipage}[t]{0.23\linewidth}
\begin{lstlisting}
@dataclass
class Wall:
    a_x: int
    a_y: int
    a_z: int
    b_x: int
    b_y: int
    b_z: int
    height: int
\end{lstlisting}
    \end{minipage} \hfill
    \begin{minipage}[t]{0.23\linewidth}
\begin{lstlisting}
@dataclass
class Door:
    wall_id: str
    position_x: int
    position_y: int
    position_z: int
    width: int
    height: int
\end{lstlisting}
    \end{minipage} \hfill
    \begin{minipage}[t]{0.23\linewidth}
\begin{lstlisting}
@dataclass
class Window:
    wall_id: str
    position_x: int
    position_y: int
    position_z: int
    width: int
    height: int
\end{lstlisting}
    \end{minipage} \hfill
    \begin{minipage}[t]{0.23\linewidth}
\begin{lstlisting}
@dataclass
class Bbox:
    <\color{black}{class}>: str
    position_x: int
    position_y: int
    position_z: int
    angle_z: int
    scale_x: int
    scale_y: int
    scale_z: int
\end{lstlisting}
    \end{minipage}
    \caption{Definition of our structured representation for layouts and objects.}
    \label{fig:parameter}
    \vspace{-3mm}
\end{figure}

Given the point cloud $\P$ (typically derived from RGBD scans), our goal is to recover the 3D layout~$\L$, which consists of all walls, doors and windows, and a set of objects $\O$ in the scene. As discussed before, we represent $\L$ and $\O$ as scripts of a general-purpose language, and fine-tune LLMs to predict the language in text form. \cref{fig:parameter} shows all the parameters we use for the layout and objects.

To the best of our knowledge, we are the first to employ LLMs for structured scene reconstruction from point clouds. Thus, we first propose a new dataset suitable for training the LLMs on (\cref{sec:method:data}). Then, with the new dataset, we set out to investigate the following two key questions: (i) What point cloud features to use for the structured indoor modeling task (\cref{sec:method:feature})? (ii) How to integrate the features into LLMs (\cref{sec:method:train})?

\subsection{Dataset and Metrics}
\label{sec:method:data}

\begin{wraptable}{r}{0.67\linewidth} 
    \vspace{-5mm}
    \centering
    \small
    \caption{Quantitative statistics of indoor RGBD scan datasets.}
    \label{tab:dataset}
    \vspace{-2mm}
    \begin{tabular}{l|cc|ccc}
        \toprule
        \multirow{2}{*}{Dataset (year)} & \multirow{2}{*}{source} & \multirow{2}{*}{\#scenes} & \multicolumn{2}{c}{annotations} \tabularnewline
        & & & layouts & objects  \tabularnewline
        \midrule
        ScanNet (2017)~\cite{ScanNet} & real & 1,513 & & $\bullet$ \tabularnewline
        Matterport3D (2017)~\cite{Matterport3D} & real & 90 &  & $\bullet$  \tabularnewline
        3RScan (2019)~\cite{WaldANTN19} & real & 1,482 &  & $\bullet$   \tabularnewline
        SceneCAD (2020)~\cite{AvetisyanKCDDN20} & real & 1,151 & $\bullet$ & $\bullet$   \tabularnewline
        ARKitScenes (2021)~\cite{ARKitScenes} & real & 5,047 &  & $\bullet$   \tabularnewline
        MultiScan (2022)~\cite{MaoZJCS22} & real & 273 &  & $\bullet$ \tabularnewline
        ScanNet++ (2023)~\cite{ScanNetpp} & real & 1,006 &   & $\bullet$  \tabularnewline
        HM3DSem (2023)~\cite{YadavRRGTGMCBSC23} & real & 181 &  & $\bullet$  \tabularnewline
        \midrule
        Structured3D (2020)~\cite{Structured3D} & syn. & 3,500 & $\bullet$ &  
        \tabularnewline
        Hypersim (2021)~\cite{Hypersim} & syn. & 461 & & $\bullet$ 
        \tabularnewline
        ProcTHOR (2022)~\cite{ProcTHOR} & syn. & 12,000 & $\bullet$ & $\bullet$ 
        \tabularnewline
        HSSD (2024)~\cite{ProcTHOR} & syn. & 211 & $\bullet$ & $\bullet$ 
        \tabularnewline
        ASE (2024)~\cite{SceneScript} & syn. & 100,000 & $\bullet$ & 
        \tabularnewline
        \midrule
        \method dataset (ours) & syn. & 12,328 & $\bullet$ & $\bullet$ \tabularnewline
        \bottomrule
    \end{tabular}
    \vspace{-3mm}
\end{wraptable}

\cref{tab:dataset} provides a list of popular indoor RGBD scan datasets. We can make a few observations:

\emph{First}, the number of scenes in real datasets is relatively small, ranging from about one hundred to a few thousands. To collect such a dataset, one needs to take RGBD scans in real scenes and manually annotate 3D information, which is a highly laborious process. Further, most real datasets provide annotations on the objects (\ie, semantic labels and 3D boxes) only. One exception is SceneCAD~\cite{AvetisyanKCDDN20}, which manually adds wall annotations to ScanNet dataset. But it only contains 1,151 rooms.

\emph{Second}, for synthetic datasets, it is much cheaper to obtain ground-truth 3D annotations. But to ensure the realism of the RGBD scans, the 3D scenes must be professionally constructed. As a result, the diversity of 3D scenes in some datasets is still limited. For example, Hypersim~\cite{Hypersim} and HSSD~\cite{HSSD} only have 461 and 211 scenes, respectively. On the other hand, one may create a large number of scenes by employing a rule-based procedural modeling pipeline, as did in ProcTHOR~\cite{ProcTHOR} and ASE~\cite{SceneScript}. But the quality of the scenes is not as high.

{\bf \method dataset}. In this paper, we make an effort to build a dataset suitable for training LLMs for structured indoor scene reconstruction. To this end, we take advantage of access to a large repository of interior designs from an online platform\footnote{\url{https://www.kujiale.com/}} in the interior design industry. Most designs are created by professional designers and used for real-world production. We parse each 3D house into individual rooms and use a few rules to filter the rooms. This results in 12,328 distinct scenes with 54,778 rooms.

For the objects, we keep annotations on a selected set of 59 common categories (excluding wall, door, and window), and filter out small objects with side lengths all  $< 15\mathrm{cm}$. In the end, our dataset contains 412,932 annotated object instances of 35,426 unique CAD models.

\begin{figure}
    \centering
    \setlength{\tabcolsep}{2pt}
    \begin{tabular}{cccc}
        \includegraphics[width=0.24\linewidth]{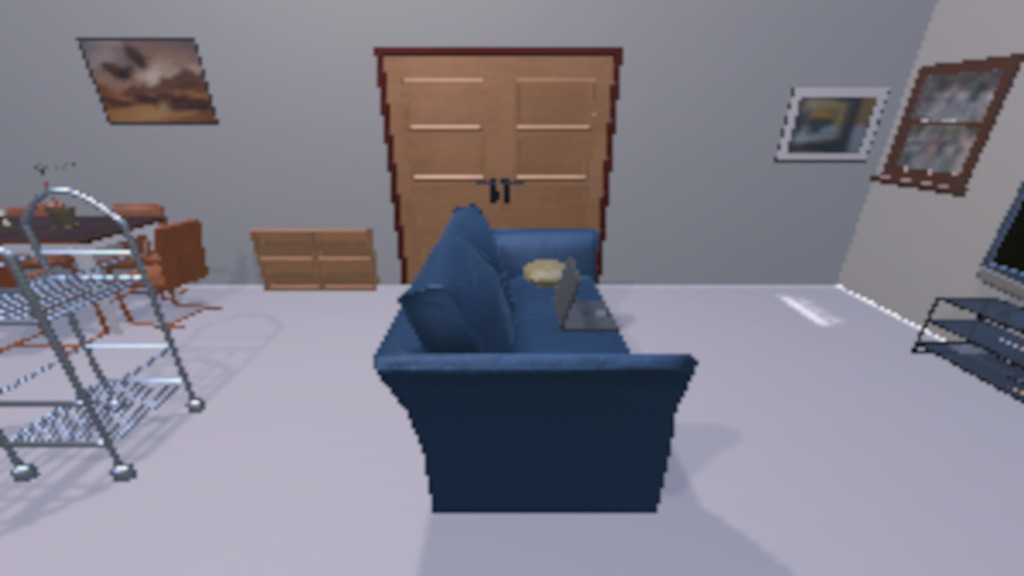} &
        \includegraphics[width=0.24\linewidth]{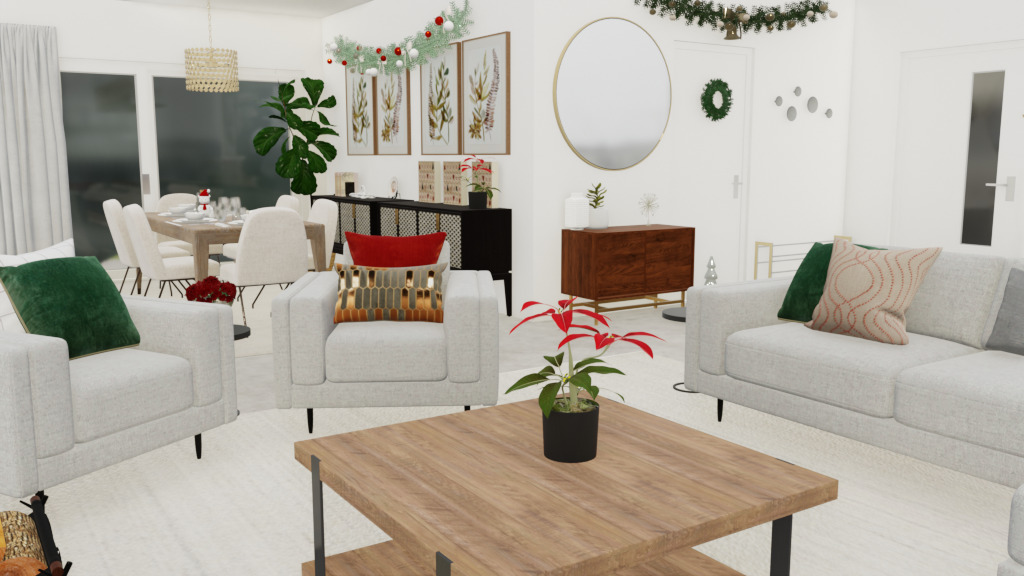} &
        \includegraphics[width=0.24\linewidth]{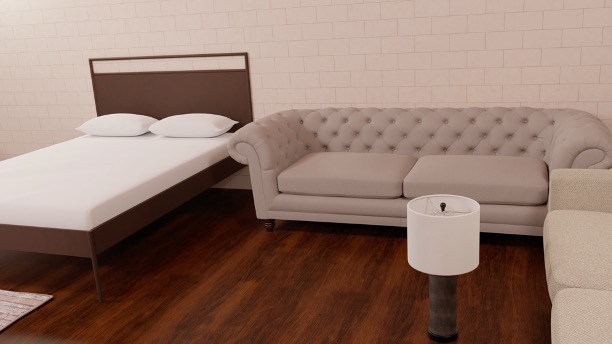} &
        \includegraphics[width=0.24\linewidth]{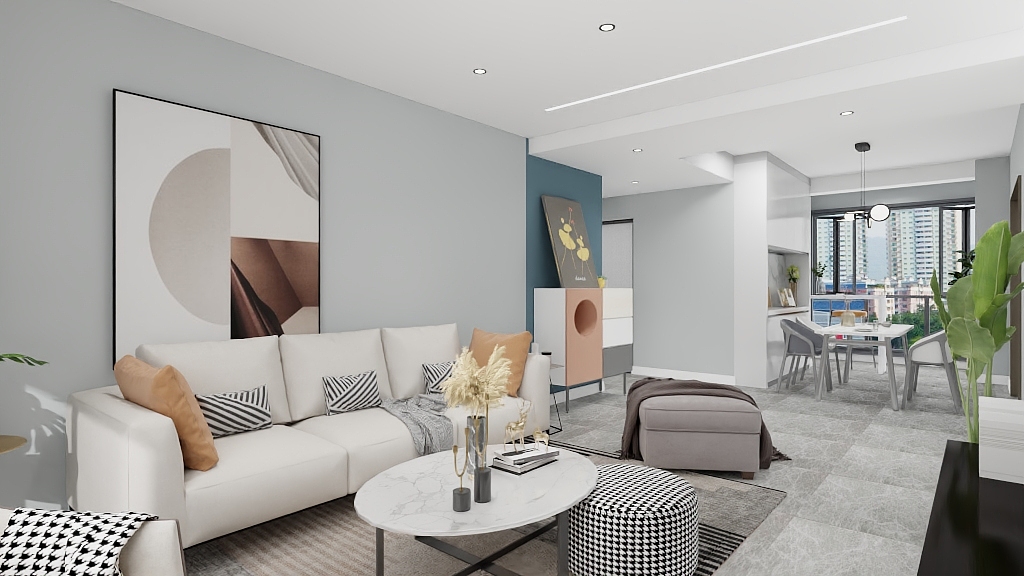} \tabularnewline
        \includegraphics[width=0.24\linewidth]{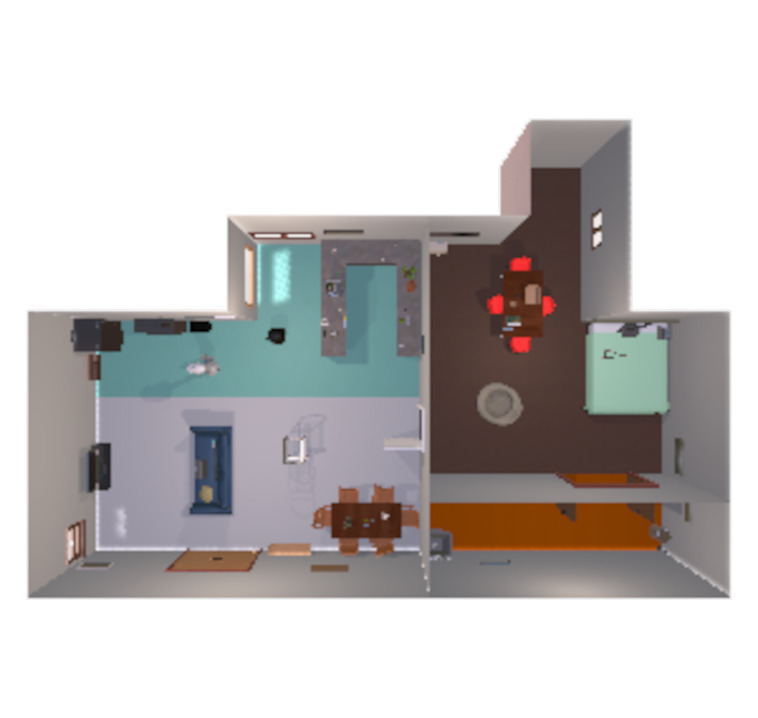} &
        \includegraphics[width=0.24\linewidth]{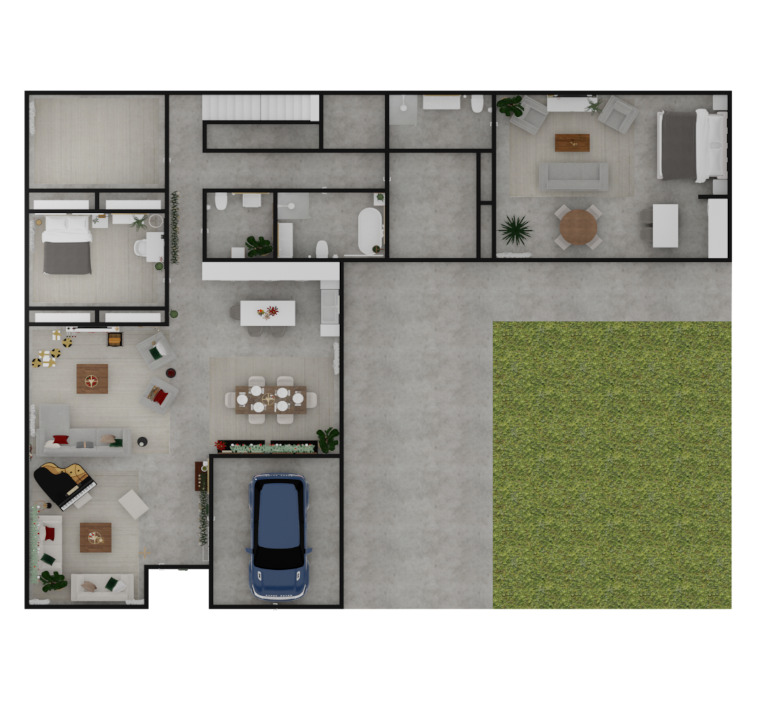} &
        \includegraphics[width=0.24\linewidth]{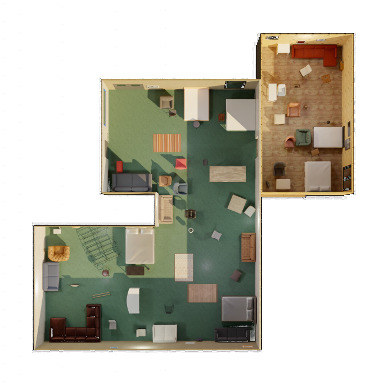} &
        \includegraphics[width=0.24\linewidth]{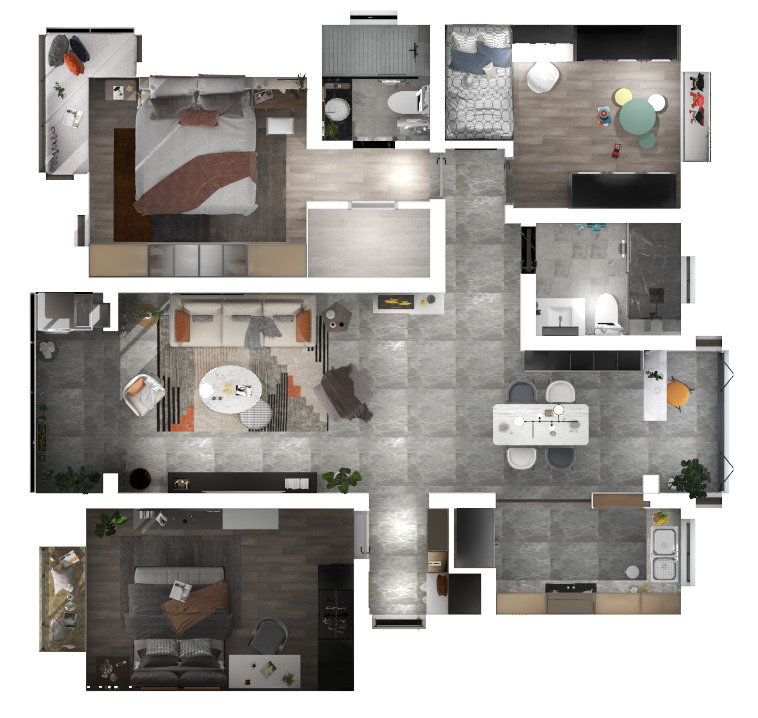} \tabularnewline
        ProcTHOR~\cite{ProcTHOR} &  HSSD~\cite{HSSD} & ASE~\cite{SceneScript} & \method dataset
    \end{tabular}
    \caption{Dataset visual quality comparison. The layout and object placements in ProcTHOR~\cite{ProcTHOR} and ASE~\cite{SceneScript} are program-generated, which exhibit noticeable differences from real-world statistics. The scenes in HSSD~\cite{HSSD} and our dataset are fully human-authored. But HSSD only has 211 scenes.}
    \label{fig:visual-quality}
\end{figure}

For each scene, we generate photo-realistic RGBD images utilizing an industry-leading rendering engine. At the time of rendering, a camera trajectory traversing each room is simulated, based on which the images are taken at an interval of $0.5\,\textrm{m}$. \cref{fig:visual-quality} compares the visual quality of our dataset to several recent synthetic datasets. Finally, the dataset is divided into 11,328/500/500 scenes for training/validation/testing.

{\bf Evaluation metrics.} We evaluate our method on both layout estimation and 3D object detection. For \emph{layout estimation}, let $\hat{\L} = \{\hat{l}_i\}_{i=1}^{\hat{m}}$ and $\L = \{l_i\}_{i=1}^{m}$ denote the set of predicted structural elements and ground truth, respectively. Since each element is a plane segment in 3D space, we compute the 2D intersection-over-union (IoU$_\textrm{2D}$) by projecting each predicted element to the plane of the ground-truth element to find the optimal assignment with the Hungarian algorithm, and report the F1-scores at 0.25 and 0.5 thresholds. For \emph{3D object detection}, we match the predictions $\hat{\O} = \{\hat{o}_i\}_{i=1}^{\hat{n}}$ with the ground truth $\O = \{o_i\}_{i=1}^{n}$ via Hungarian matching. Then, we compute the 3D intersection-over-union (IoU$_\textrm{3D}$) between each matched pair and report the F1-scores at 0.25 and 0.5 thresholds.

\subsection{Point Cloud Encoders}
\label{sec:method:feature}

Extracting meaningful representations for a point cloud is challenging because of its irregular format. Learning-based approaches to process 3D point clouds can be roughly classified into three groups: \emph{Mapping-based methods} are inspired by the success of 2D image backbones such as CLIP~\cite{CLIP} and the DINO series~\cite{DINO, DINOv2}, and map pixel-aligned image features to 3D. For example, Lexicon3D~\cite{Lexicon3D} projects 3D points onto the image plane to obtain their corresponding pixel features; ConceptFusion~\cite{JatavallabhulaK23} builds multimodal 3D maps via traditional SLAM and multi-view fusion approaches. \emph{Voxel-based methods} first transform irregular point clouds to regular representations via 3D voxelization. Then, sparse convolution algorithms~\cite{ChoyGS19} can be readily used for efficient computation. \emph{Point-based methods}, by contrast, process point clouds directly as sets embedded in continuous space. Recent approaches in this category~\cite{ZhaoJJTK21, 0002LJLZ22, PTv3, Sonata} have successfully explored the self-attention mechanism in Transformers and demonstrated superior performance on downstream applications.

Note that all methods discussed above produce one feature per point by default. Given that a scene-level point cloud consists of hundreds of thousands of points or more, these features are too much in terms of quantity for LLMs to consume directly. Since most modern open-source LLMs take an ``Encoder-MLP-LLM'' architecture for multimodal feature alignment (\eg, the Llama series~\cite{llama3} and Qwen series~\cite{qwen2.5,qwen2.5-vl}), we can define a point cloud encoder for LLMs as a mapping which takes a set of $N$ points as input and generates $K$ feature embeddings, where $K\ll N$:
\begin{equation}
    \F = \mathcal{E}(\P), \quad \P \in \mathbb{R}^{N\times 6}, \F \in \mathbb{R}^{K\times D}.
\label{eqn:1}
\end{equation}
Here, each input point is a 6-dimensional vector (XYZ and RGB), and $D$ is the feature dimension.

Recently, a few attempts have been made to convert point representations into feature embeddings~\cite{SceneScript, Scene-LLM}. However, this step has not been carefully investigated. In this section, we conduct experiments to study (i) the effectiveness of different encoders, and (ii) the impact of $K$, \ie, the number of output features. Unless otherwise stated, Qwen2.5-0.5B~\cite{qwen2.5} is employed as our base model. We use Sonata~\cite{Sonata} as the point cloud encoder, and a two-layer MLP as the projector.

{\bf Experiment on the encoder design.} We consider six possible point cloud encoders. The first group of encoders adopts a mapping-based method to fuse 2D image features to 3D points. Following Lexicon3D~\cite{Lexicon3D}, we pool pixel-level DINOv2 features to form 3D point representations. Then, one of the two routes may be taken to reduce the points into a manageable length of tokens:
\begin{itemize}
    \item {\bf Voxelize~($\P$+DINOv2)}: divides the space into a fixed-resolution voxel grid and aggregates all the points in the same voxel -- similar to the approach used in Scene-LLM~\cite{Scene-LLM};
    \item {\bf Rand.~sampling~($\P$+DINOv2)}: uses Farthest Point Sampling~\cite{QiYSG17} to randomly sample a fixed number of points with their features.
\end{itemize}
From \cref{tab:encoder} we can see, these two options result in extremely low F1-scores. We hypothesize that this is because too much spatial information is lost during the naive down-sampling step. While this may be acceptable for certain applications, such as visual Q\&A, it is problematic for our task, which relies heavily on the spatial information to reconstruct the geometric coordinates of objects.

\begin{table}[t]
    \centering
    \small
    \setlength{\tabcolsep}{2.5pt}
    \caption{Experiment on the encoder design.}
    \label{tab:encoder}
    \begin{tabular}{l|c|cc|cc}
        \toprule
        \multirow{2}{*}{Encoder} & Params & \multicolumn{2}{c|}{F1 - layout} & \multicolumn{2}{c}{F1 - object} \tabularnewline
        & (M) & IoU$_\textrm{2D}$@0.25 & IoU$_\textrm{2D}$@0.5 & IoU$_\textrm{3D}$@0.25 & IoU$_\textrm{3D}$@0.5 \tabularnewline
        \midrule
        Voxelize~($\P$+DINOv2) & 0.0 & 10.8 & 5.5 & 0.1 & 0.0 \tabularnewline
        Rand.~sampling~($\P$+DINOv2) & 0.0 & 10.6 & 4.9 & 0.2 & 0.0\tabularnewline
        \midrule
        3DCNN enc.~($\P$) & 10.5 & 79.4 & 76.3 & 59.8 & 46.1 \tabularnewline
        3DCNN enc.~($\P$+DINOv2) & 11.6  & 81.5 & 79.4 & 57.1  & 45.0 \tabularnewline
        3DCNN enc.~($\P$) +  Voxelize~(DINOv2) & 11.9 & 83.8 & 81.6 & 62.9 & 46.7 \tabularnewline
        \midrule
        Sonata/PTv3 enc.~($\P$) & 108.8 & {\bf 84.6} & {\bf 82.6} & {\bf 65.1} & {\bf 49.4} \tabularnewline
        \bottomrule
    \end{tabular}
    \vspace{-3mm}
\end{table}

The second group of encoders adopts a standard 3DCNN encoder to generate pooled features. We follow SceneScript~\cite{SceneScript} to build our encoder with a sparse 3D convolution library~\cite{TangLL0022}. Here we consider three variants:
\begin{itemize}
    \item {\bf 3DCNN enc.~($\P$)}: takes raw points (6-dim XYZRGB values) as input;
    \item {\bf 3DCNN enc.~($\P$+DINOv2)}: takes the points with the associated DINOv2 features as input;
    \item {\bf 3DCNN enc.~($\P$) +  Voxelize~(DINOv2)}: concatenates the output of 3DCNN with aggregated DINOv2 features.
\end{itemize}
As shown in \cref{tab:encoder}, a learned encoder is effective in keeping the necessary semantic and geometric information for reconstruction. We also find that incorporating DINOv2 features yields slightly better results, possibly due to the enhanced contextual information. 

The last encoder we consider is {\bf Sonata}~\cite{Sonata}, a variant of Point Transformer V3 (PTv3)~\cite{PTv3} which removes the decoder and focuses on self-supervised learning on the encoder. Compared to other backbones, such an ``encoder-only'' design makes it quite convenient for adapting to LLMs. Indeed, as shown in \cref{tab:encoder}, with raw points as input, it already achieves the best results among all encoders.

\begin{wraptable}{r}{0.67\linewidth} 
    \centering
    \small
    \setlength{\tabcolsep}{2.5pt}
    \caption{Experiment on the spatial resolution. Avg.~($K$) indicates the average value of $K$ under the corresponding resolution.}
    \label{tab:resolution}
    \begin{tabular}{l|c|cc|cc}
        \toprule
        \multirow{2}{*}{Res.} & \multirow{2}{*}{Avg.~($K$)} & \multicolumn{2}{c|}{F1 - layout} & \multicolumn{2}{c}{F1 - object} \tabularnewline
        & & IoU$_\textrm{2D}$@0.25 & IoU$_\textrm{2D}$@0.5 & IoU$_\textrm{3D}$@0.25 & IoU$_\textrm{3D}$@0.5 \tabularnewline
        \midrule
        1$\times$ & 123 & 84.6 & 82.6 & 65.1 & 49.4 \tabularnewline
        1.25$\times$ & 197 & 84.8 & 82.6 & 65.3 & 52.2 \tabularnewline
        1.5$\times$ & 286 & 86.9 & 85.1 & 68.7 & 58.6 \tabularnewline
        2$\times$ & 510 & {\bf 87.4} & {\bf 85.3} & 71.6 & 61.1 \tabularnewline
        2.5$\times$ & 788 & 86.1 & 83.9 & {\bf 72.2} & {\bf 61.7} \tabularnewline
        4$\times$ & 2,000 & 85.8 & 83.7 & 69.9 & 58.9 \tabularnewline
        \bottomrule
    \end{tabular}
\end{wraptable}

\textbf{Experiment on the spatial resolution}. Recall in \cref{eqn:1} that the parameter $K$ denotes the number of feature embeddings or visual tokens, which is critical to the performance of LLMs. In this experiment, we study the effect of $K$.

However, fixing $K$ is difficult because different point clouds generally produce varying $K$ values. Instead, since the encoders employ a coarse-to-fine structure for point cloud processing, we set the resolution at the finest level at $5\mathrm{cm}$ and use a 5-level hierarchical structure by default. A detailed illustration of the point cloud encoder hierarchy is presented in the appendix (see \cref{fig:sonata-hierarchy}). Then, we gradually increase the resolution up to 4$\times$, and report the results in \cref{tab:resolution}. As can be seen,  model performance steadily improves from 1$\times$ to 2$\times$ resolutions, but further increasing the resolution hurts the performance. This is likely due to the excessively long token sequences.

\subsection{Training Schedules}
\label{sec:method:train}

\begin{wraptable}{r}{0.67\linewidth} 
    \centering
    \small
    \setlength{\tabcolsep}{2.5pt}
    \caption{Experiment on training schedule. For each training stage, we use three circles to represent the three components in the ``Encoder-MLP-LLM'' architecture. ``$\circ$'' and ``$\bullet$'' indicate whether a component is frozen or trainable, respectively.}
    \label{tab:train}
    \begin{tabular}{ccc|cc|cc}
        \toprule
        \multicolumn{3}{c|}{Training stages} & \multicolumn{2}{c|}{F1 - layout} & \multicolumn{2}{c}{F1 - object} \tabularnewline
        1st & 2nd & 3rd & IoU$_\textrm{2D}$@0.25 & IoU$_\textrm{2D}$@0.5 & IoU$_\textrm{3D}$@0.25 & IoU$_\textrm{3D}$@0.5 \tabularnewline
        \midrule
        $\circ\bullet\bullet$ & - & - & 73.8 & 68.4 & 40.2 & 22.7\tabularnewline
        $\bullet\bullet\bullet$ & - & - & 84.6 & 82.6 & {\bf 65.1} & {\bf 49.4} \tabularnewline
        \midrule
        $\bullet\bullet\circ$ & $\circ\bullet\bullet$ & - & 70.9 & 62.7 & 27.4 & 13.7 \tabularnewline
        $\bullet\bullet\circ$ & $\bullet\bullet\bullet$ & - & {\bf 85.3} & {\bf 83.4} & 60.7 & 45.6\tabularnewline
        \midrule
        $\circ\bullet\circ$ & $\bullet\bullet\circ$ & $\circ\bullet\bullet$ & 76.9 & 70.1 & 31.1 & 16.0 \tabularnewline
        $\circ\bullet\circ$ & $\bullet\bullet\circ$ & $\bullet\bullet\bullet$  & 83.2 & 80.3 & 57.1 & 41.2 \tabularnewline
        \bottomrule
    \end{tabular}
\end{wraptable}

As most recent MLLMs start from pre-trained unimodal backbones, how to align the multimodal content with the language model plays a critical role in the performance of the resulting MLLMs. For example, a common practice adopted by many MLLMs~\cite{LiuLWL23a, TongBWWIAYYMWPF24} is the inclusion of a two-stage training schedule: (i) an alignment stage that trains the randomly initialized MLP projector only, followed by (ii) a fine-tuning stage that trains both the projector and language model while keeping the vision encoder frozen. Other studies (\eg,~\cite{Karamcheti0BLKS24}) advocate a single-stage training protocol, suggesting that including the projector training stage is unnecessary. Furthermore, when adapting the LLMs to rich visual content such as long videos, some research~\cite{apollo} has shown that a three-stage pipeline, which progressively unfreezes different components in different stages, improves the performance.

In this experiment, we test six possible training configurations, ranging from one-stage to three-stage, to evaluate the influence of different training strategies. As shown in \cref{tab:train}, we found that a single-stage fine-tuning yields the best results. Adding more stages generally degrades the model performance. We also found that making all parameters trainable in a single stage is important, especially for 3D object detection. This may indicate that, compared to modern image encoders such as SigLIP~\cite{ZhaiM0B23} and DINOv2~\cite{DINOv2}, current pre-trained point cloud encoders are not as versatile in supporting downstream tasks yet.

\section{Experiments}
\label{sec:exp}

In this section, we compare \method to the current state-of-the-art on scene layout estimation (\cref{sec:exp:layout}) and 3D object detection (\cref{sec:exp:object}), respectively. Following our empirical study, we use Qwen2.5-0.5B~\cite{qwen2.5} as the base model and Sonata~\cite{Sonata} as the point cloud encoder. We set of resolution at the finest level at $2.5\mathrm{cm}$ and employ a single-stage training on our dataset.

Besides testing on established RGBD benchmarks, we also investigate whether our model can handle inputs from other sources, such as point clouds reconstructed from RGB videos (\cref{sec:exp:video}). Other extensions and more technical details can be found in the appendix.

\subsection{Layout Estimation}
\label{sec:exp:layout}

Layout estimation focuses on predicting architectural elements, \ie, walls, doors, and windows, within an indoor scene. In this experiment, we include results from two representative baselines, namely RoomFormer~\cite{YueKSE23} and SceneScript~\cite{SceneScript}. As the state-of-the-art specialist model for layout estimation, RoomFormer employs a highly specialized network architecture, using two-level queries to predict room polygons and their corners, respectively. SceneScript, on the other hand, casts 3D structure estimation as an auto-regressive ``next-token-prediction'' problem. However, unlike our method, it uses specialized structured language commands to describe the scene and trains a Transformer decoder from scratch. 

Following the setting of RoomFormer, we perform evaluation on the Structured3D benchmark~\cite{Structured3D}, which contains 3,500 residential houses with diverse floorplans. We use the original data split of 3000/250/250 for training/validation/testing, respectively. For RoomFormer, we directly download the model checkpoint from its website\footnote{\url{https://github.com/ywyue/RoomFormer}} for inference. Note that RoomFormer takes in 2D density maps projected from point clouds and outputs 2D layouts, which are converted to 3D by extruding based on the lowest/highest points in the point cloud. For SceneScript, we download the pre-trained model and code\footnote{\url{https://github.com/facebookresearch/scenescript}}, and fine-tune it on Structured3D with the default hyperparameter settings.

\begin{wraptable}{r}{0.72\linewidth}
    \centering
    \small
    \setlength{\tabcolsep}{2.5pt}
    \caption{Experiment on layout estimation.}
    \label{tab:layout}
    \begin{tabular}{l|c|cc}
        \toprule
        \multirow{2}{*}{Method} & Params & \multicolumn{2}{c}{F1 - Structured3D} \tabularnewline
        & (M) & IoU$_\textrm{2D}$@0.25 & IoU$_\textrm{2D}$@0.5 \tabularnewline
        \midrule
        RoomFormer~\cite{YueKSE23} & 42.2  & 83.4 & 81.4  \tabularnewline
        SceneScript~\cite{SceneScript} & 29.9  & 90.4  & 89.2 \tabularnewline
        \midrule
        \method (ft.~Structured3D) & 603.5 & 32.6 & 18.1 \tabularnewline
        \method (ft.~Ours) & 603.5 & 59.9 & 44.7 \tabularnewline
        \method (ft.~Ours $\rightarrow$ Structured3D) & 603.5 & {\bf 94.3} & {\bf 93.5} \tabularnewline
        \bottomrule
    \end{tabular}
\end{wraptable}
For our model, we include three variants: (i) trained on Structured3D only, (ii) trained on our dataset only, and (iii) first trained on our dataset, then fine-tuned on Structured3D. As can be seen in \cref{tab:layout}, directly training LLMs on the smaller Structured3D dataset leads to poor performance. In contrast, by first training on our new dataset and then on Structured3D, our model outperforms both baselines for this task. This demonstrates the benefits of our new dataset. 

\begin{figure}[t]
    \centering
    \setlength{\tabcolsep}{2pt}
    \begin{tabular}{cccc}
        \includegraphics[width=0.24\linewidth]{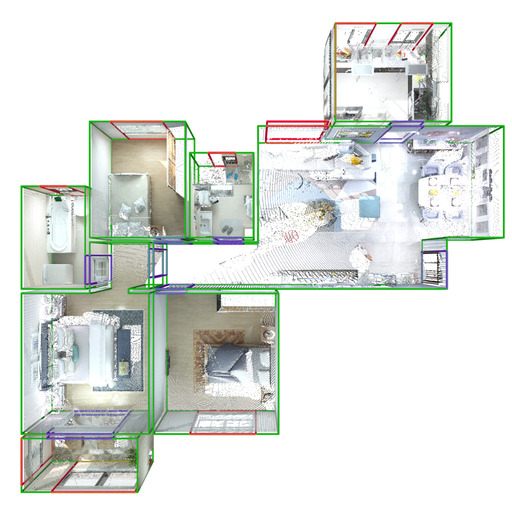} &
        \includegraphics[width=0.24\linewidth]{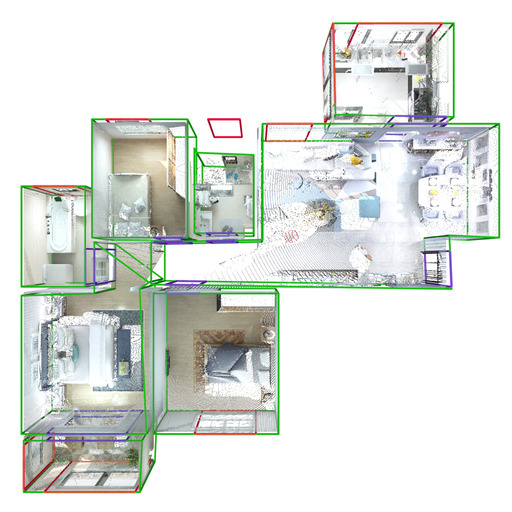} &
        \includegraphics[width=0.24\linewidth]{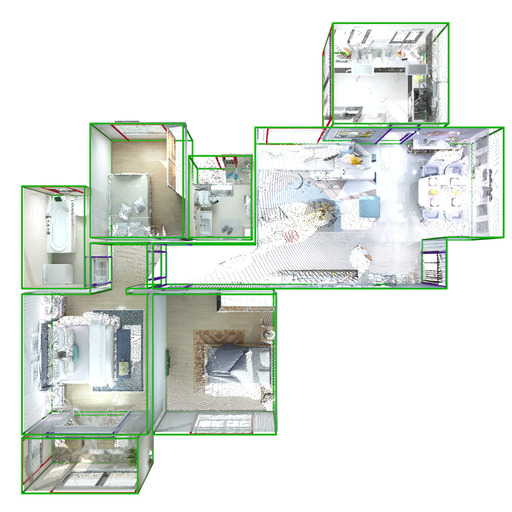} &
        \includegraphics[width=0.24\linewidth]{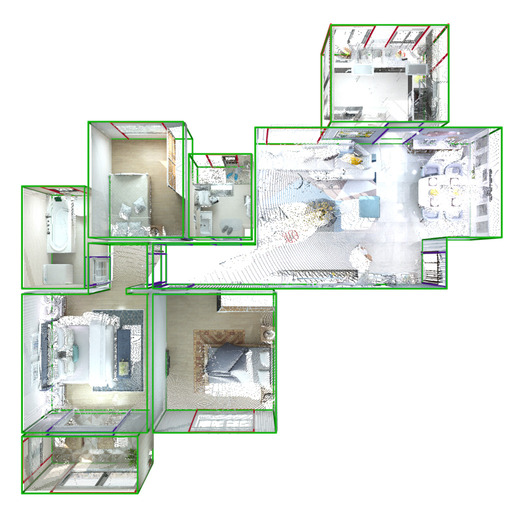} \tabularnewline

        \includegraphics[width=0.24\linewidth]{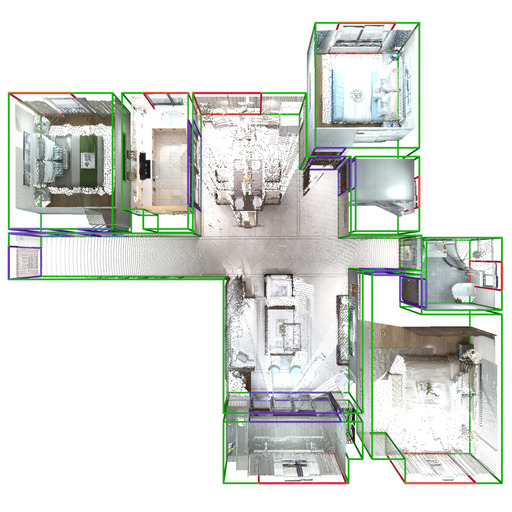} &
        \includegraphics[width=0.24\linewidth]{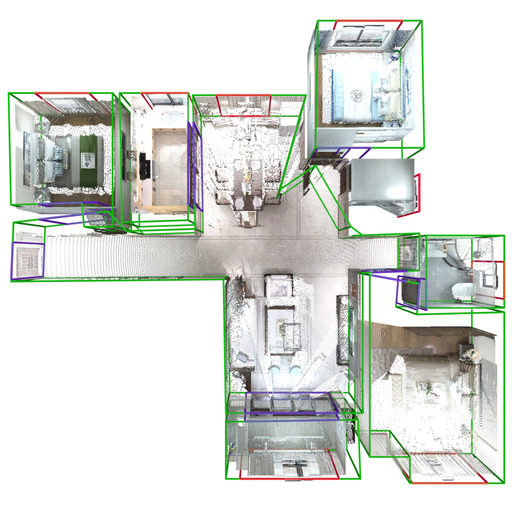} &
        \includegraphics[width=0.24\linewidth]{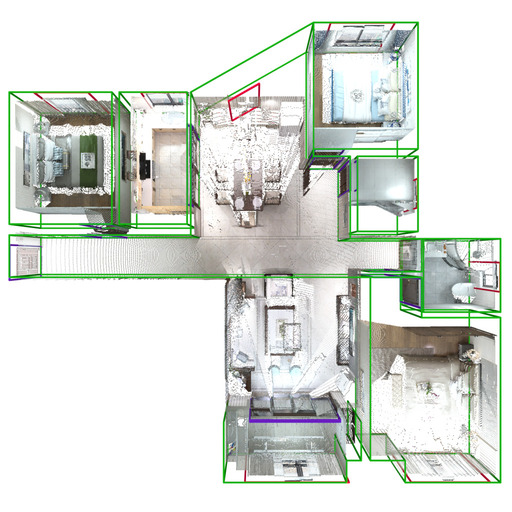} &
        \includegraphics[width=0.24\linewidth]{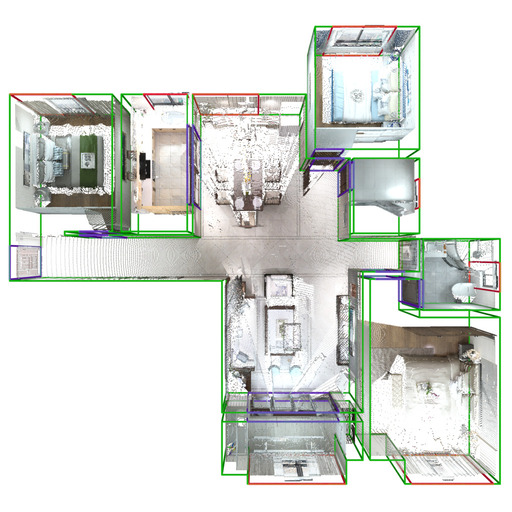} \tabularnewline
        
        Ground truth & RoomFormer~\cite{YueKSE23} & SceneScript~\cite{SceneScript} & \method
    \end{tabular}
    \caption{Qualitative results on Structured3D dataset.}
    \label{fig:stru3d}
\end{figure}

\cref{fig:stru3d} provides some qualitative comparisons. Since RoomFormer detects room elements independently, we can see that some doors and windows are not attached to any wall. Auto-regressive methods like SceneScript and our method can preserve the relationships between structures by design. But we observe that SceneScript tends to miss some elements in its output.

\subsection{3D Object Detection}
\label{sec:exp:object}

Next, we compare \method to existing methods on 3D object detection. We include results from two representative baselines, namely V-DETR~\cite{0001GYLLW00G24} and SceneScript. As the state-of-the-art specialist model for the task, V-DETR builds upon the DETR framework~\cite{DETR} and makes several task-specific improvements, including a 3D vertex relative position encoding (3DV-RPE), object-based normalization, among others. Following prior practice, we perform evaluation on ScanNet~\cite{ScanNet}, which contains 1,513 3D indoor scans with annotations of 18 object categories. The training set is composed of 1,201 scenes, while 312 scenes are used for validation/testing. Since auto-regressive models (\ie, \method and SceneScript) do not produce confidence scores, we report F1 scores instead of mean Average Precision (mAP), as suggested in~\cite{SceneScript}.

\begin{wraptable}{r}{0.66\linewidth} 
    \centering
    \small
    \setlength{\tabcolsep}{2.5pt}
    \caption{Experiment on 3D object detection. ``*'': numbers reported on the overlapping subset of object classes between our dataset and ScanNet.}
    \label{tab:object}
    \begin{tabular}{l|c|cc}
        \toprule
        \multirow{2}{*}{Method} & Params & \multicolumn{2}{c}{F1 - ScanNet} \tabularnewline
        & (M) & IoU$_\textrm{3D}$@0.25 & IoU$_\textrm{3D}$@0.5 \tabularnewline
        \midrule
        V-DETR~\cite{0001GYLLW00G24} & 79.4 & 65.1 & {\bf 56.8}  \tabularnewline
        SceneScript~\cite{SceneScript} & 29.9 & 49.1 & 36.8 \tabularnewline
        \midrule
        \method (ft.~ScanNet) & 603.5 & 2.9 & 0.7 \tabularnewline
        \method (ft.~Ours) & 603.5 & 33.8$^*$ & 22.6$^*$ \tabularnewline
        \method (ft.~Ours $\rightarrow$ ScanNet) & 603.5 & \textbf{65.6} & 52.6 \tabularnewline
        \bottomrule
    \end{tabular}
\end{wraptable}
For V-DETR, we directly use the model checkpoint downloaded from its website\footnote{\url{https://github.com/V-DETR/V-DETR}}. A confidence threshold of 0.5 is applied for both objectness and semantic predictions, followed by non-maximum suppression (NMS). For SceneScript, we fine-tune the model on ScanNet with the default hyperparameter settings. During fine-tuning, we apply the same data augmentations as used in PTv3~\cite{PTv3} and V-DETR~\cite{0001GYLLW00G24}.

For our model, we again include three variants: (i) trained on ScanNet only, (ii) trained on our dataset only, and (iii) first trained on our dataset, then fine-tuned on ScanNet. As can be seen in \cref{tab:object}, the first variant has very low F1 scores, suggesting that ScanNet itself is too small to train LLMs on. By first training on our new dataset and then on ScanNet, our model outperforms SceneScript and achieves competitive results against the state-of-the-art specialist V-DETR.

\begin{figure}[t]
    \centering
    \setlength{\tabcolsep}{2pt}
    \begin{tabular}{cccc}
        \includegraphics[width=0.24\linewidth]{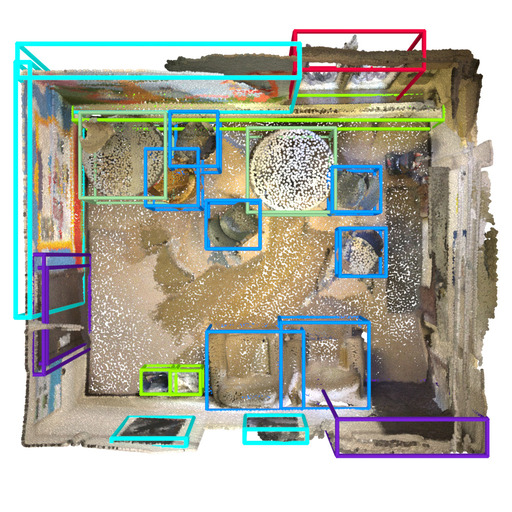} &
        \includegraphics[width=0.24\linewidth]{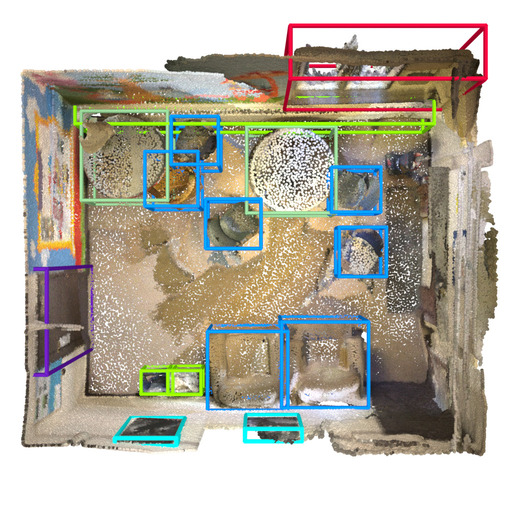} &
        \includegraphics[width=0.24\linewidth]{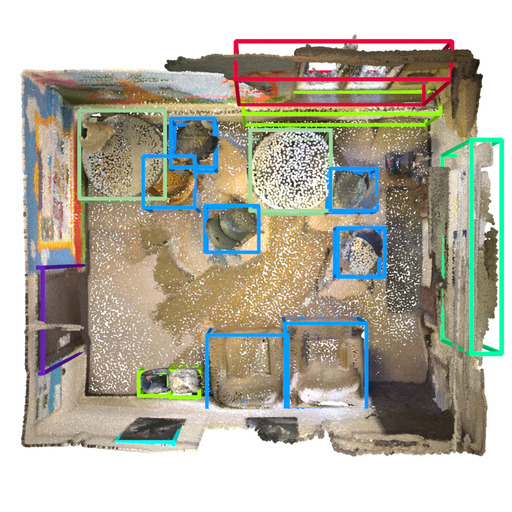} &
        \includegraphics[width=0.24\linewidth]{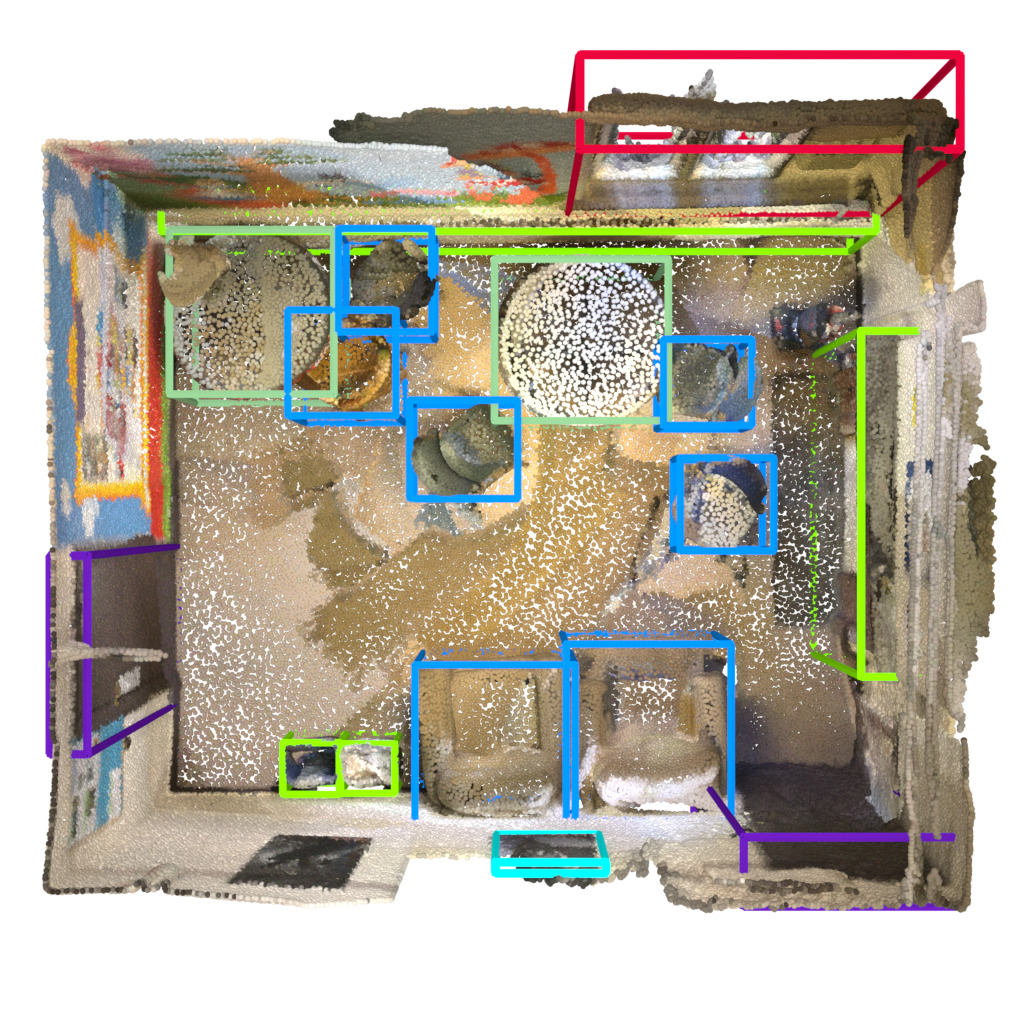} \tabularnewline
        \includegraphics[width=0.24\linewidth]{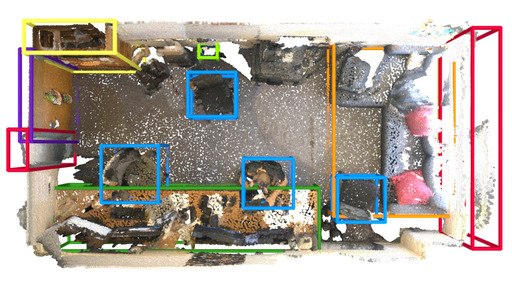} &
        \includegraphics[width=0.24\linewidth]{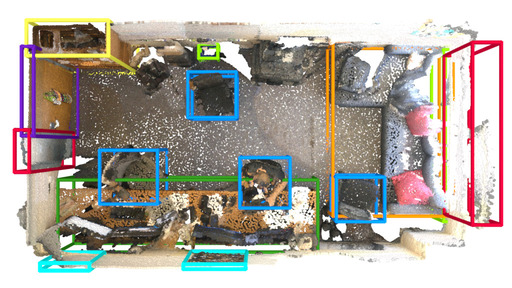} &
        \includegraphics[width=0.24\linewidth]{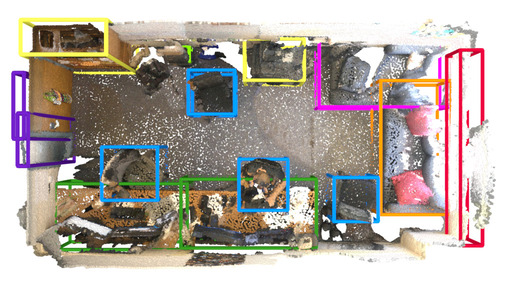} &
        \includegraphics[width=0.24\linewidth]{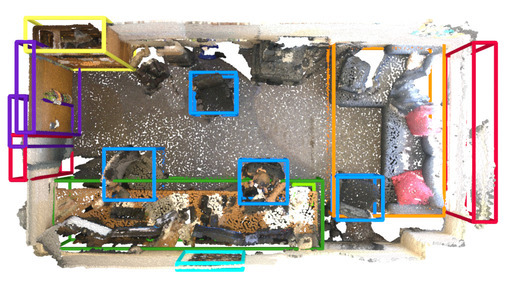} \tabularnewline
        Ground truth & V-DETR~\cite{0001GYLLW00G24} & SceneScript~\cite{SceneScript} & \method
    \end{tabular}
    \caption{Qualitative results on ScanNet dataset.}
    \label{fig:scannet}
\end{figure}

\cref{fig:scannet} shows some qualitative results. Compared to SceneScript, \method demonstrates improved detection accuracies across categories. In comparison to V-DETR, we observe that our method has some difficulty in locating objects in ``picture'', ``sink'' and ``showercurtain'' classes. Notably, ``picture'' and ``sink'' are the smallest objects in ScanNet by volume ($< 0.1\mathrm{m^3}$), whereas ``showercurtain'' category is not included in our dataset and has the fewest occurrences in the ScanNet.

\subsection{Zero-shot Detection on Videos}
\label{sec:exp:video}

Although significant progress has been made in SFM and SLAM techniques over the past decades, reliably inferring 3D structure from videos remains a highly challenging problem. Recently, DUSt3R~\cite{DUSt3R} and its extensions~\cite{MASt3R, MASt3R-SfM, MAST3RSLAM, slam3r} provide a novel solution that employs deep networks for simultaneous point matching and camera motion estimation. These techniques are shown to be remarkably robust in dealing with real-world videos or unconstrained image collections.

In this experiment, we collect a set of 107 videos of virtual indoor scene tours and use MASt3R-SLAM~\cite{MAST3RSLAM} to reconstruct a point cloud from each video. Then, we test \method in a zero-shot setting on the reconstructed point clouds. Unlike RGBD scans, these reconstructions have significant noises, occlusions, and geometric artifacts, posing a greater challenge for structured 3D understanding. 

\begin{figure}[t]
    \centering
    \setlength{\tabcolsep}{2pt}
    \begin{tabular}{c|ccc}
        \includegraphics[width=0.25\linewidth,valign=m]{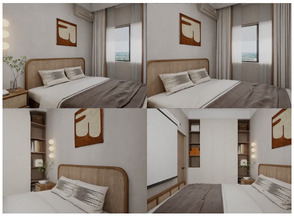} &
        \includegraphics[width=0.3\linewidth,valign=m]{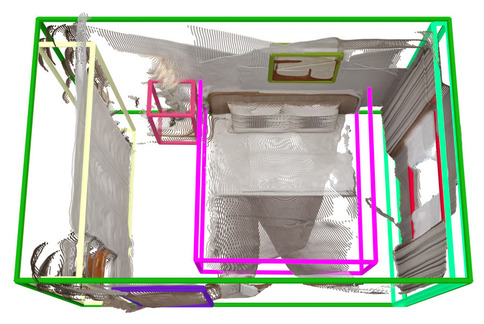} &
        \includegraphics[width=0.2\linewidth,valign=m]{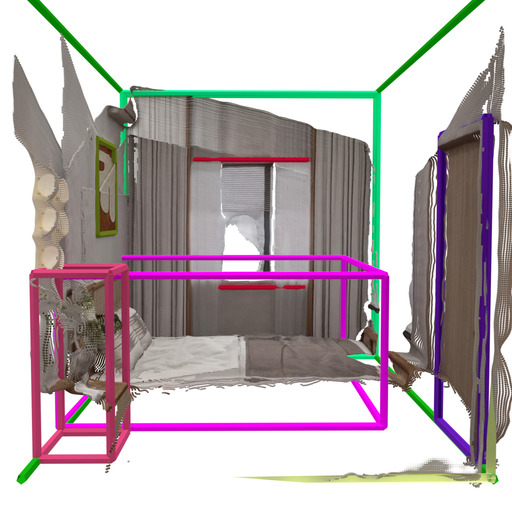} &
        \includegraphics[width=0.2\linewidth,valign=m]{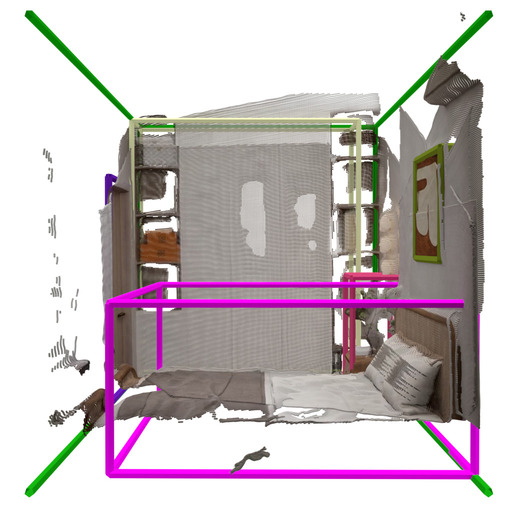} \vspace{-3mm} \tabularnewline
        \includegraphics[width=0.25\linewidth,valign=m]{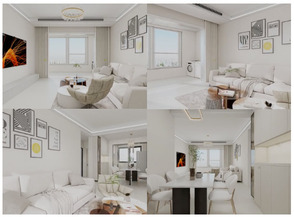} &
        \includegraphics[width=0.3\linewidth,valign=m]{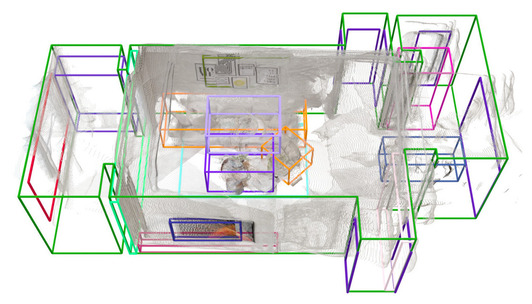} &
        \includegraphics[width=0.2\linewidth,valign=m]{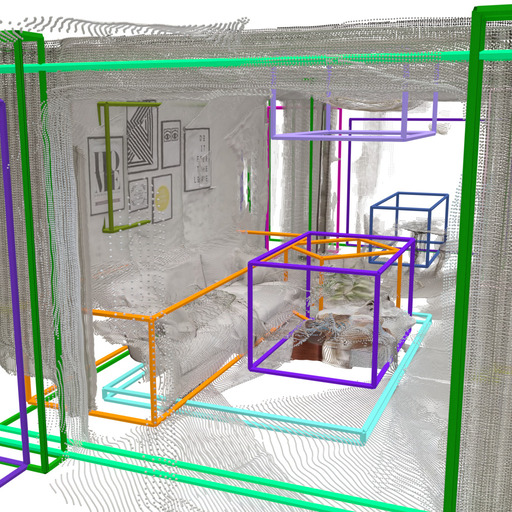} &
        \includegraphics[width=0.2\linewidth,valign=m]{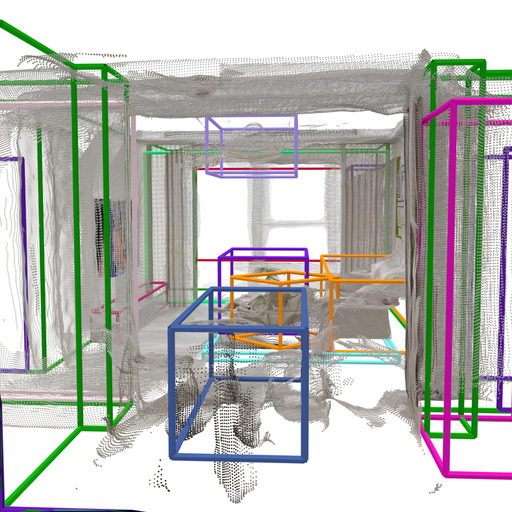} \tabularnewline
    \end{tabular}
    \caption{Qualitative results of zero-shot detection on videos.}
    \label{fig:zeroshot}
\end{figure}

\cref{fig:zeroshot} shows some qualitative results. Without fine-tuning, \method demonstrates strong resilience to imperfect data and preserves consistent layout and object-level predictions. Trained on synthetic datasets with accurate bounding boxes, the model infers full object sizes and orientations from sparse or occluded inputs. In the first example, it predicts beds and nightstands extending to the floor, even when those regions are not visible. Leveraging the auto-regressive nature of LLMs, \method is also capable of generating plausible room layouts from partial observations. In the second example, the model reasonably reconstructs areas such as the balcony and dining space, filling in missing elements based on contextual understanding.

\section{Conclusion}
\label{sec:conclusion}

In this study, we demonstrate the feasibility of training LLMs for structured indoor modeling tasks. We regard it as a meaningful step towards building future foundation models that can not only understand, but also reason about, interact with, and even create structured 3D scenes.

{\bf Limitations.} \method has several limitations. {\em First}, it falls short of delivering a universal, state-of-the-art model for extracting 3D structure from {\em arbitrary} point clouds. There are significant differences between point clouds from different sources, such as monocular videos, RGBD scans, and LiDAR sensors. As shown in \cref{sec:exp}, although our model exhibits reasonable generalizability across datasets, fine-tuning on a specific dataset is still needed to achieve the best performance. Further scaling up the data and model size could be a promising future direction. {\em Second}, in this paper, we focus on training LLM for structured indoor modeling only, without considering how it affects the model's natural skills in natural language processing and reasoning. To address this, a more thorough benchmarking may be performed. {\em Third}, our current approach models indoor layouts using a set of predefined object categories, limiting the LLMs in leveraging their open-ended language capabilities. Future directions include extending our work to support open-vocabulary object detection and 3D visual question answering (VQA), enabling more flexible and generalizable scene understanding. 

{\bf Impacts.} We envision that \method will be used in real-world applications such as layout estimation and 3D object detection. For example, people may take our code and fine-tune the model on their own datasets. On the methodology side, we hope that our work will inspire future studies on MLLM models for 3D scene understanding, reasoning, and creation.

\begin{ack}
This work was partially supported by the Key R\&D Program of Zhejiang Province (2025C01001) and the HKUST project 24251090T019.
\end{ack}

{
\bibliographystyle{plainnat}
\bibliography{bibliography}
}

\newpage
\appendix

\newpage
\section{\method Dataset}
\label{sec:supp:dataset}

\method dataset comprises 12,328 distinct scenes with 54,778 rooms, featuring a total of 403,291 walls, 123,301 doors, and 48,887 windows, with an overall floor area of approximately 863,986 $\mathrm{m^2}$.

\begin{figure}[t]
    \centering
    \includegraphics[width=\linewidth]{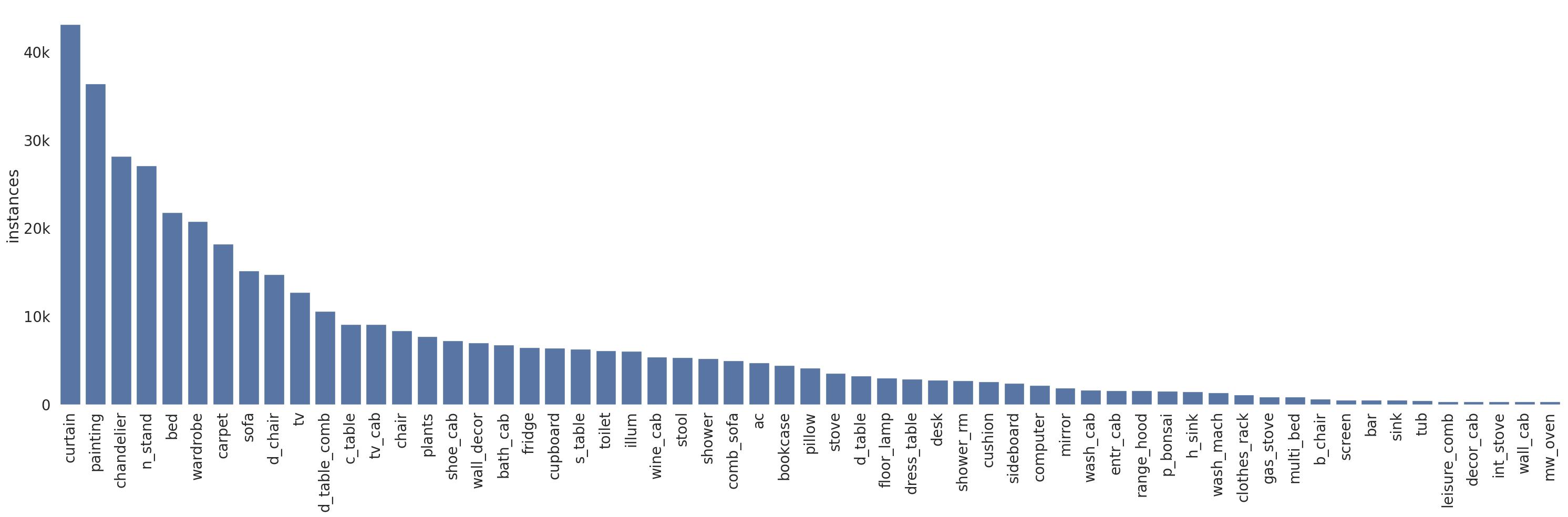} \tabularnewline
    \includegraphics[width=\linewidth]{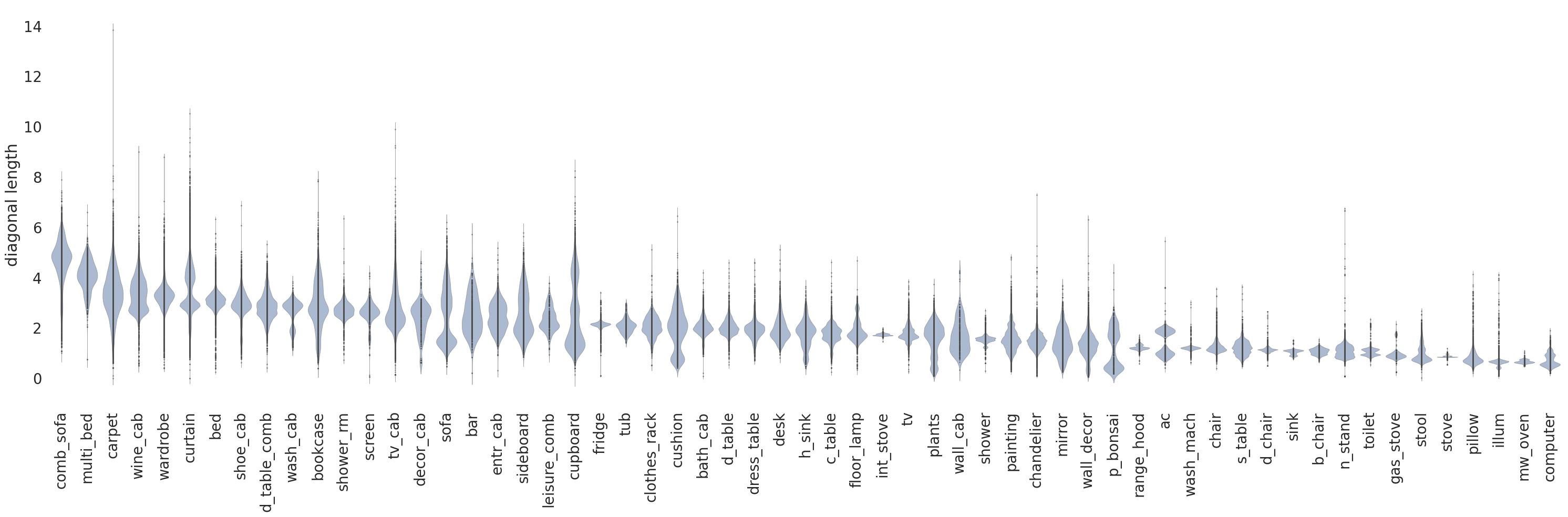}
    \caption{Statistics of the \method dataset. {\bf Top:} Number of objects in each category. {\bf Bottom:} Distribution of physical sizes in each category. We compute the object size as the diagonal length of the object's bounding box (in meters). All objects in our dataset have realistic physical sizes.}
    \label{fig:object-stats}
\end{figure}

{\bf Object categories.} We have curated a set of 59 commonly occurring object categories, organized by functional and semantic similarity, as shown in \cref{tab:categories}. Statistics of these categories can be found in \cref{fig:object-stats}.

\begin{table}[ht]
    \centering
    \small
    \setlength{\tabcolsep}{4pt}
    \renewcommand{\arraystretch}{1.3}
    \caption{Categories in the \method dataset.}
    \begin{tabular}{@{}l>{\raggedright\arraybackslash}p{0.7\linewidth}@{}}
        \toprule
        \textbf{Super Categories} & \textbf{Categories} \tabularnewline
        \midrule
        Seatings & sofa, chair, dining\_chair, bar\_chair, stool \tabularnewline
        Beddings & bed, pillow \tabularnewline
        Cabinetry & wardrobe, nightstand, tv\_cabinet, wine\_cabinet, bathroom\_cabinet, shoe\_cabinet, entrance\_cabinet, decorative\_cabinet, washing\_cabinet, wall\_cabinet, sideboard, cupboard \tabularnewline
        Tables & coffee\_table, dining\_table, side\_table, dressing\_table, desk \tabularnewline
        Kitchen appliances & integrated\_stove, gas\_stove, range\_hood, microwave\_oven, sink, stove, refrigerator \tabularnewline
        Bathroom fixtures & hand\_sink, shower, shower\_room, toilet, tub \tabularnewline
        Lighting & illumination, chandelier, floor\_standing\_lamp \tabularnewline
        Decoration & wall\_decoration, painting, curtain, carpet, plants, potted\_bonsai \tabularnewline
        Electronics & tv, computer, air\_conditioner, washing\_machine \tabularnewline
        Others & clothes\_rack, mirror, bookcase, cushion, bar, screen, combination\_sofa, dining\_table\_combination, leisure\_table\_and\_chair\_combination, multifunctional\_combination\_bed \tabularnewline
        \bottomrule
    \end{tabular}
    \label{tab:categories}
\end{table}

\begin{figure}[t]
    \centering
    \includegraphics[width=\linewidth]{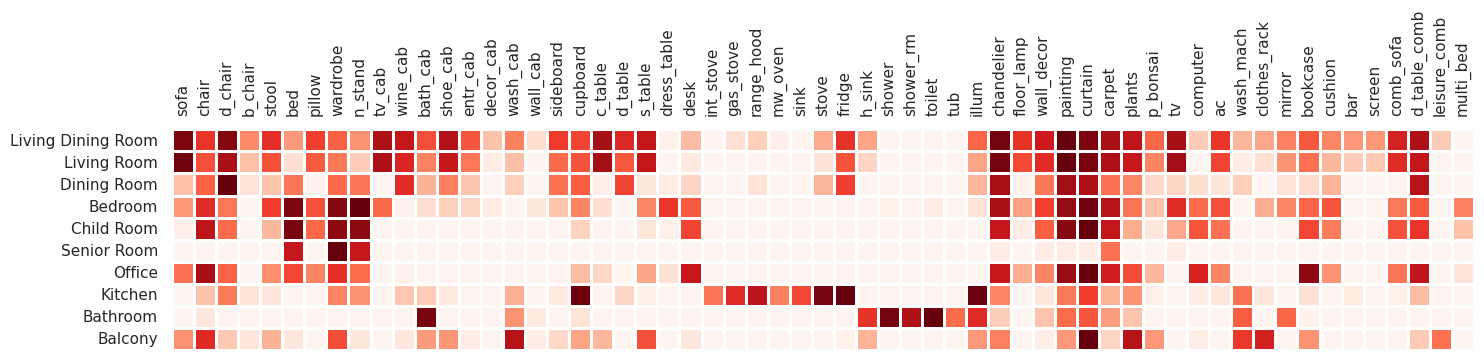}
    \caption{Room object correlation of \method dataset.}
    \label{fig:room-object-correlation}
\end{figure}

\cref{fig:room-object-correlation} plots the object-to-room correlation heatmap, which reveals strong correlations between object distribution and specific room types. Additionally, each room type contains a variety of object categories, highlighting the diversity of our dataset.

\begin{figure}[ht]
    \centering
    \setlength{\tabcolsep}{2pt}
    \renewcommand{\arraystretch}{4}
    \begin{tabular}{ccccc}
        \rotatebox{90}{\hspace{-2em}{living room}} &  
        \includegraphics[width=0.2\linewidth,valign=m]{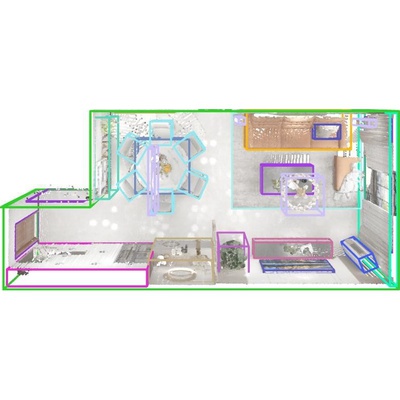} &
        \includegraphics[width=0.2\linewidth,valign=m]{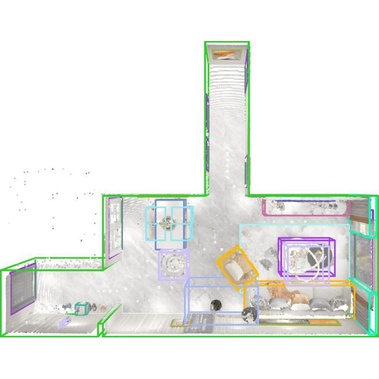} &
        \includegraphics[width=0.2\linewidth,valign=m]{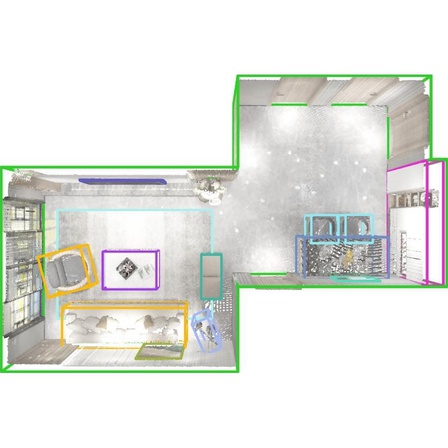} &
        \includegraphics[width=0.2\linewidth,valign=m]{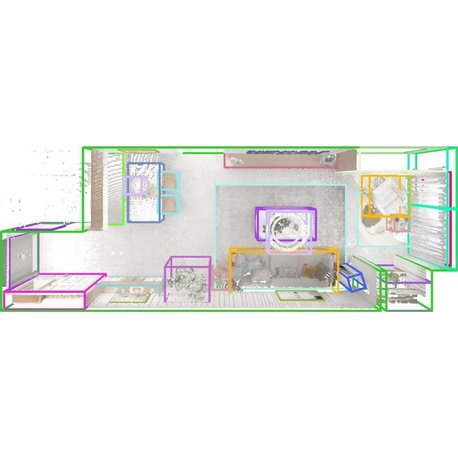} \tabularnewline

        \rotatebox{90}{\hspace{-2em}{living room}} &  
        \includegraphics[width=0.2\linewidth,valign=m]{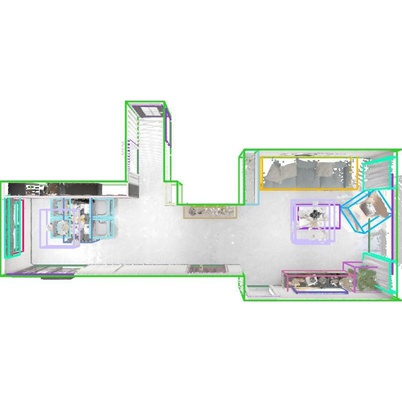} &
        \includegraphics[width=0.2\linewidth,valign=m]{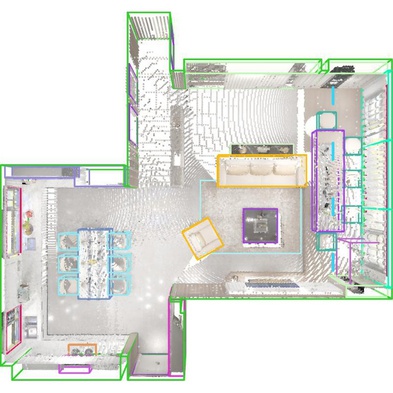} &
        \includegraphics[width=0.2\linewidth,valign=m]{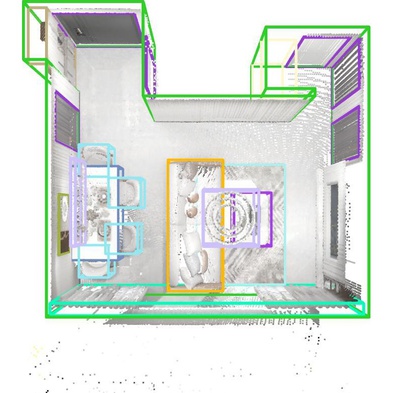} &
        \includegraphics[width=0.2\linewidth,valign=m]{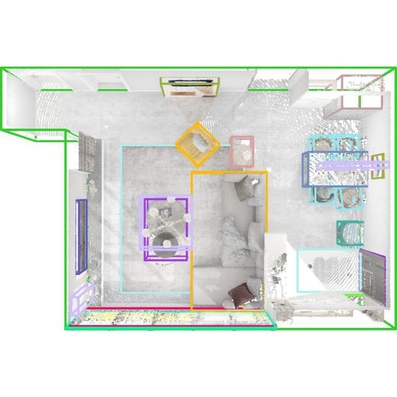} \tabularnewline

        \rotatebox{90}{\hspace{-2em}{living room}} &  
        \includegraphics[width=0.2\linewidth,valign=m]{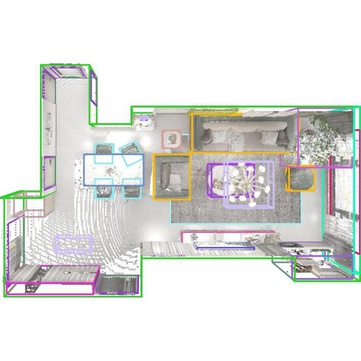} &
        \includegraphics[width=0.2\linewidth,valign=m]{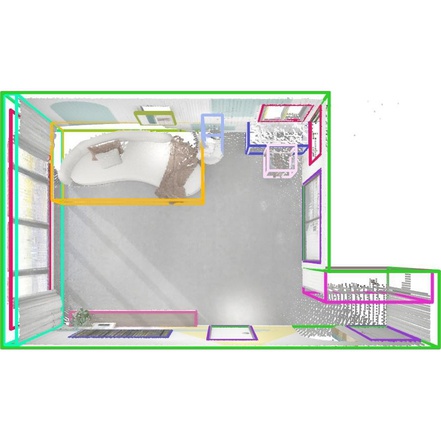} &
        \includegraphics[width=0.2\linewidth,valign=m]{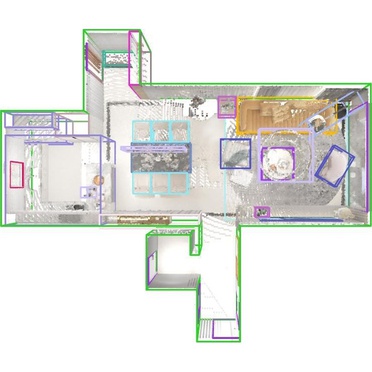} &
        \includegraphics[width=0.2\linewidth,valign=m]{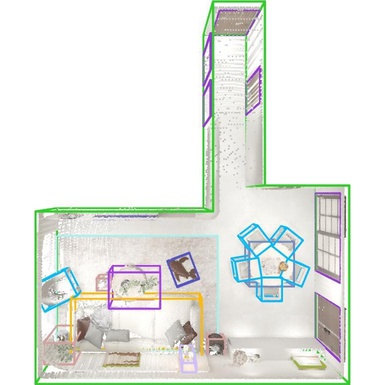} \tabularnewline

        \rotatebox{90}{\hspace{-2em}{living room}} &  
        \includegraphics[width=0.2\linewidth,valign=m]{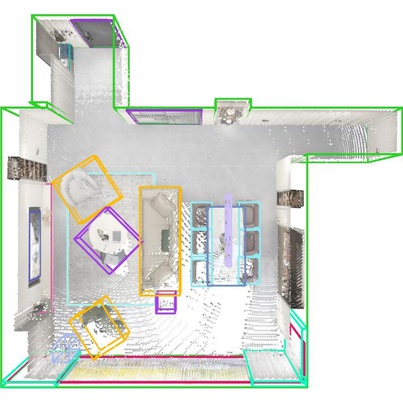} &
        \includegraphics[width=0.2\linewidth,valign=m]{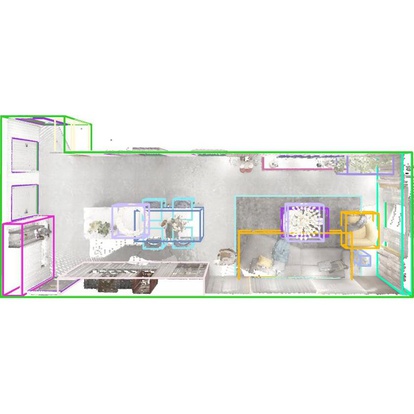} &
        \includegraphics[width=0.2\linewidth,valign=m]{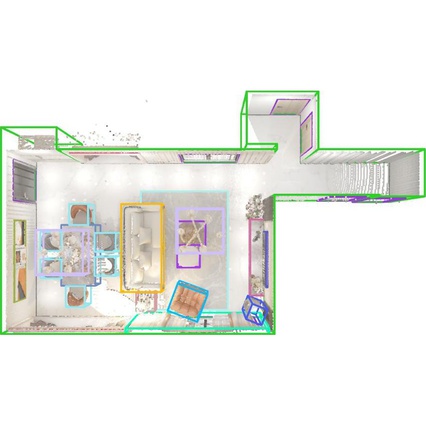} &
        \includegraphics[width=0.2\linewidth,valign=m]{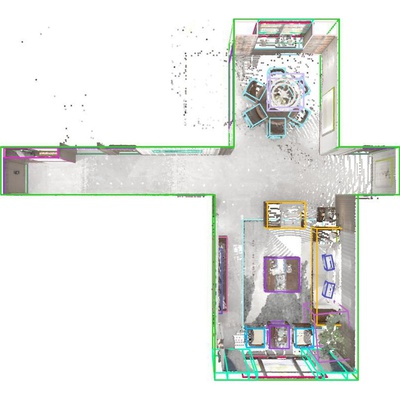} \tabularnewline

        \rotatebox{90}{\hspace{-1.5em}{bedroom}} &  
        \includegraphics[width=0.2\linewidth,valign=m]{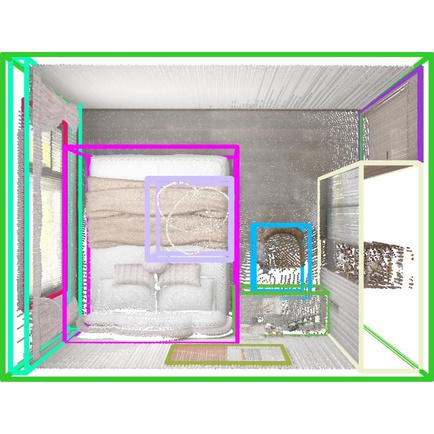} &
        \includegraphics[width=0.2\linewidth,valign=m]{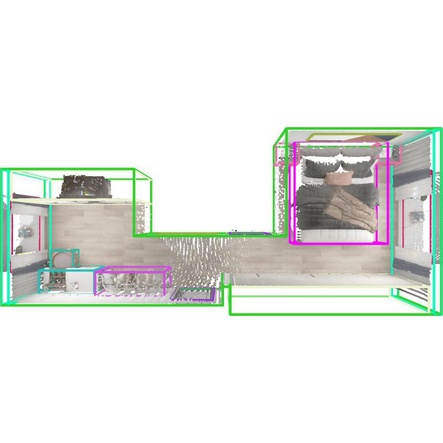} &
        \includegraphics[width=0.2\linewidth,valign=m]{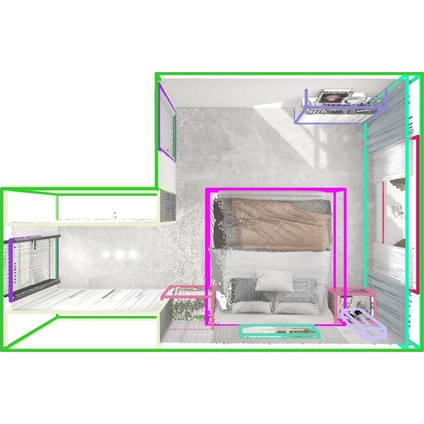} &
        \includegraphics[width=0.2\linewidth,valign=m]{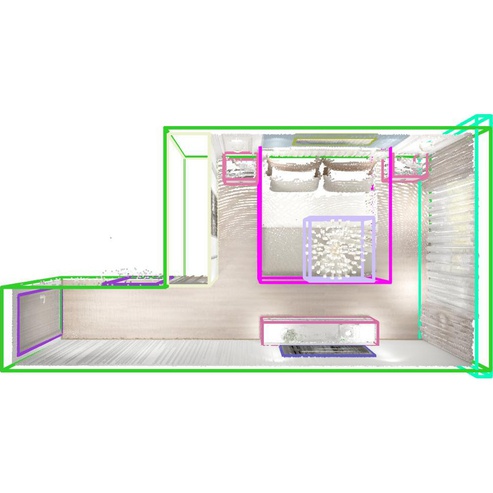} \tabularnewline

        \rotatebox{90}{\hspace{-1.5em}{bedroom}} &  
        \includegraphics[width=0.2\linewidth,valign=m]{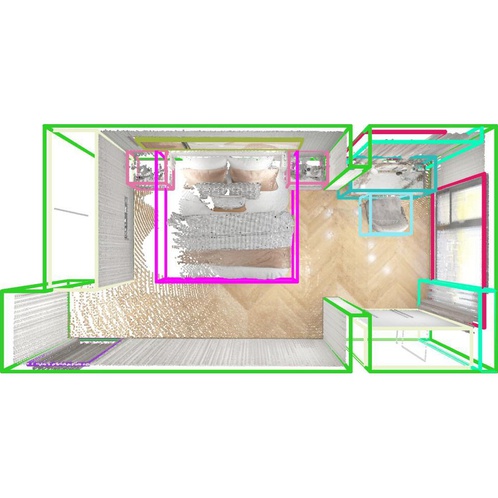} &
        \includegraphics[width=0.2\linewidth,valign=m]{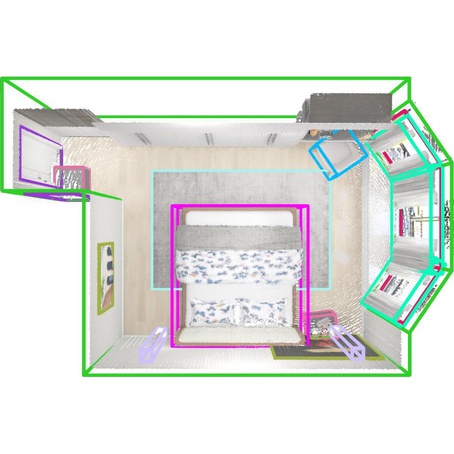} &
        \includegraphics[width=0.2\linewidth,valign=m]{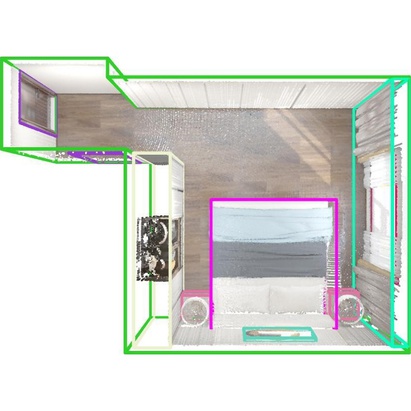} &
        \includegraphics[width=0.2\linewidth,valign=m]{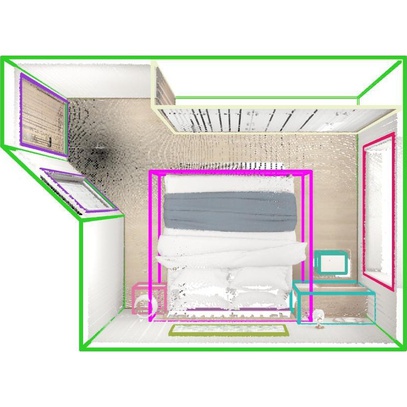} \tabularnewline

        \rotatebox{90}{\hspace{-1.5em}{bedroom}} &  
        \includegraphics[width=0.2\linewidth,valign=m]{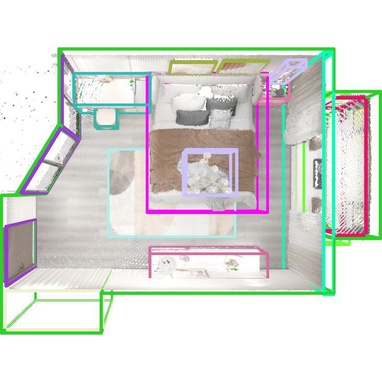} &
        \includegraphics[width=0.2\linewidth,valign=m]{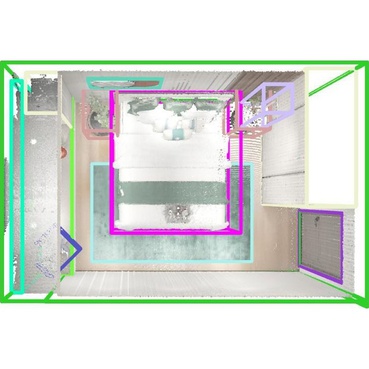} &
        \includegraphics[width=0.2\linewidth,valign=m]{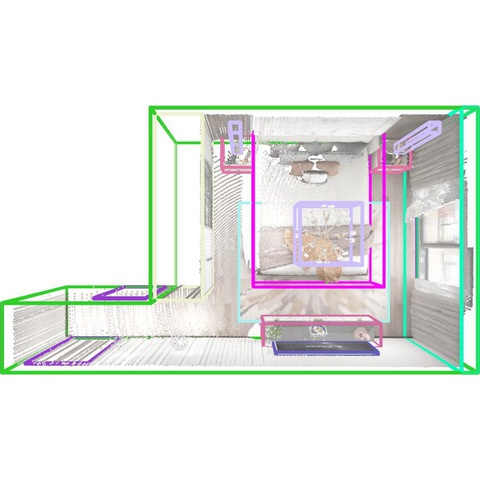} &
        \includegraphics[width=0.2\linewidth,valign=m]{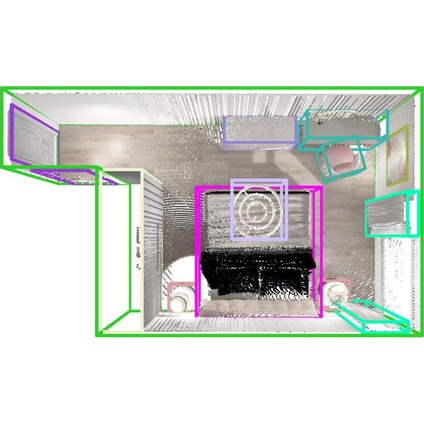} \tabularnewline
        
        \rotatebox{90}{\hspace{-1.5em}{bathroom}} &  
        \includegraphics[width=0.2\linewidth,valign=m]{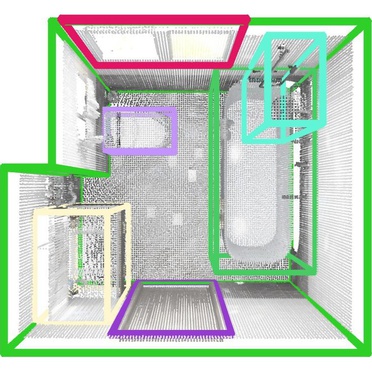} &
        \includegraphics[width=0.2\linewidth,valign=m]{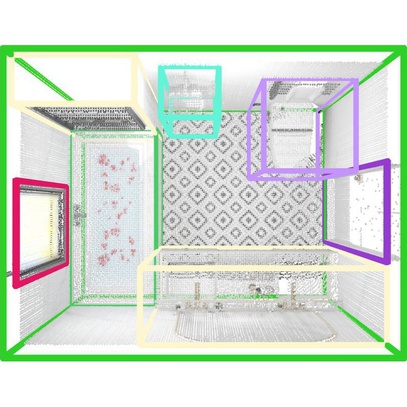} &
        \includegraphics[width=0.2\linewidth,valign=m]{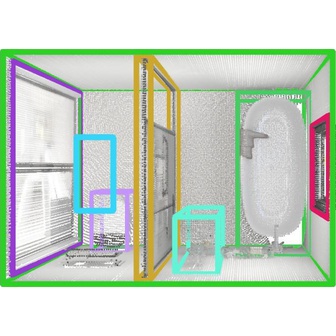} &
        \includegraphics[width=0.2\linewidth,valign=m]{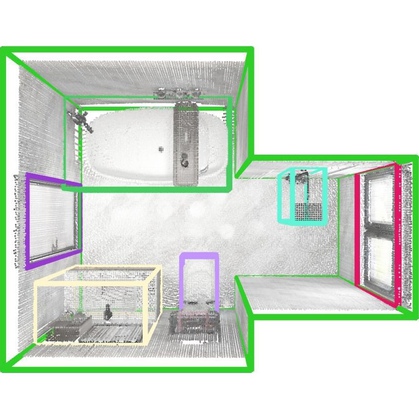} \tabularnewline
    \end{tabular}
    \caption{Point clouds with overlaid 3D layout and object annotations in \method dataset.}
    \label{fig:visual-layout}
\end{figure}

\begin{figure}[ht]
    \centering
    \setlength{\tabcolsep}{2pt}
    \renewcommand{\arraystretch}{4}
    \begin{tabular}{ccccccccc}
        sofa &
        \includegraphics[width=0.1\linewidth,valign=m]{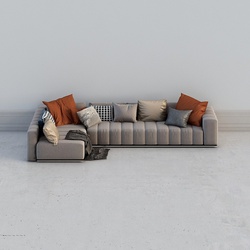} &
        \includegraphics[width=0.1\linewidth,valign=m]{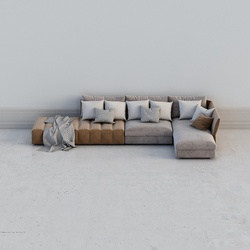} &
        \includegraphics[width=0.1\linewidth,valign=m]{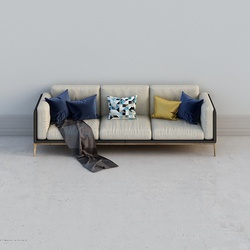} &
        \includegraphics[width=0.1\linewidth,valign=m]{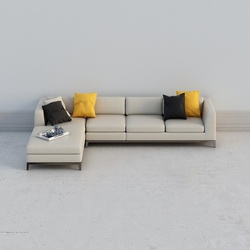} &
        \includegraphics[width=0.1\linewidth,valign=m]{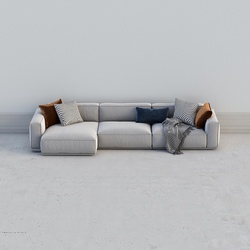} &
        \includegraphics[width=0.1\linewidth,valign=m]{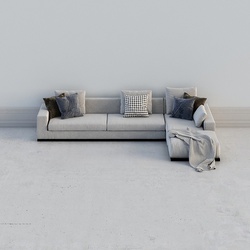} &
        \includegraphics[width=0.1\linewidth,valign=m]{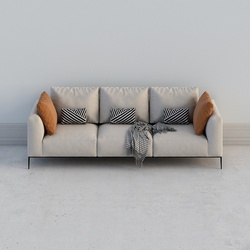} &
        \includegraphics[width=0.1\linewidth,valign=m]{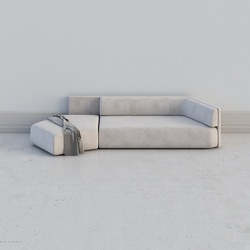} \tabularnewline
        coffee table &
        \includegraphics[width=0.1\linewidth,valign=m]{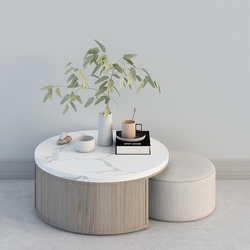} &
        \includegraphics[width=0.1\linewidth,valign=m]{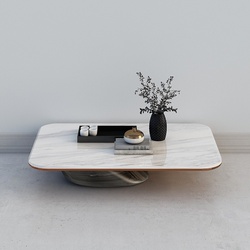} &
        \includegraphics[width=0.1\linewidth,valign=m]{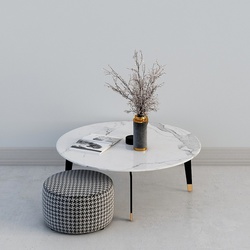} &
        \includegraphics[width=0.1\linewidth,valign=m]{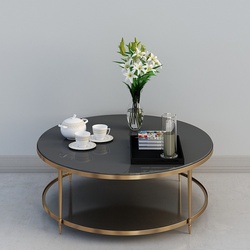} &
        \includegraphics[width=0.1\linewidth,valign=m]{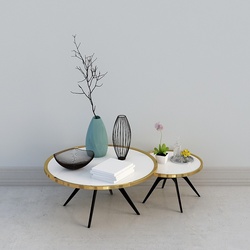} &
        \includegraphics[width=0.1\linewidth,valign=m]{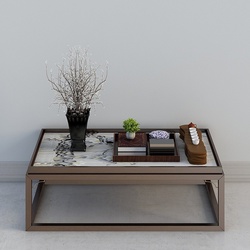} &
        \includegraphics[width=0.1\linewidth,valign=m]{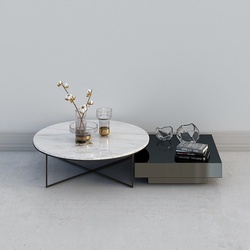} &
        \includegraphics[width=0.1\linewidth,valign=m]{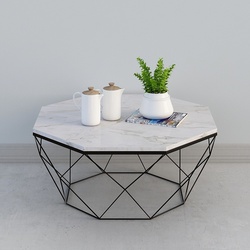} \tabularnewline
        tv cabinet &
        \includegraphics[width=0.1\linewidth,valign=m]{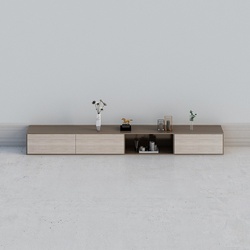} &
        \includegraphics[width=0.1\linewidth,valign=m]{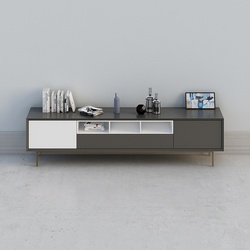} &
        \includegraphics[width=0.1\linewidth,valign=m]{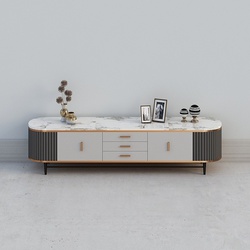} &
        \includegraphics[width=0.1\linewidth,valign=m]{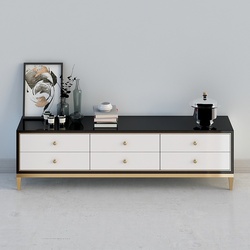} &
        \includegraphics[width=0.1\linewidth,valign=m]{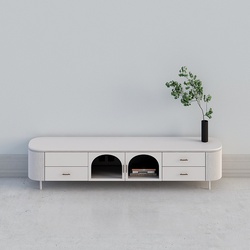} &
        \includegraphics[width=0.1\linewidth,valign=m]{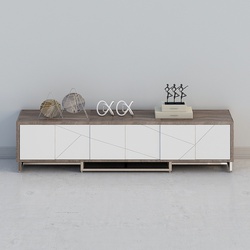} &
        \includegraphics[width=0.1\linewidth,valign=m]{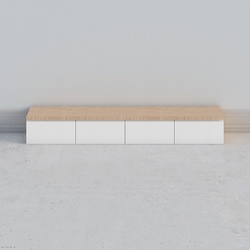} &
        \includegraphics[width=0.1\linewidth,valign=m]{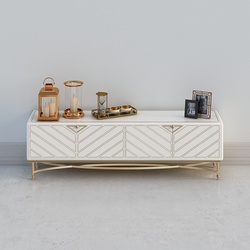} \tabularnewline
        bed &
        \includegraphics[width=0.1\linewidth,valign=m]{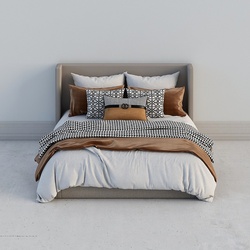} &
        \includegraphics[width=0.1\linewidth,valign=m]{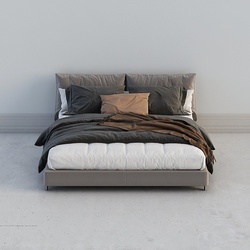} &
        \includegraphics[width=0.1\linewidth,valign=m]{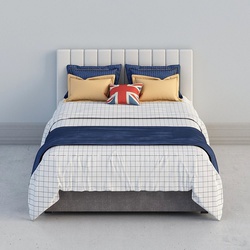} &
        \includegraphics[width=0.1\linewidth,valign=m]{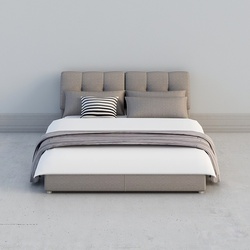} &
        \includegraphics[width=0.1\linewidth,valign=m]{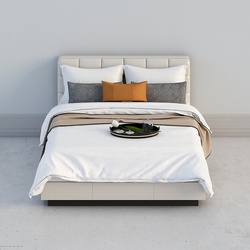} &
        \includegraphics[width=0.1\linewidth,valign=m]{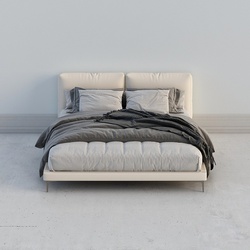} &
        \includegraphics[width=0.1\linewidth,valign=m]{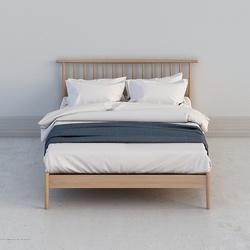} &
        \includegraphics[width=0.1\linewidth,valign=m]{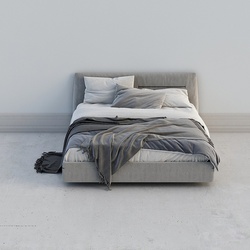} \tabularnewline
        nightstand &
        \includegraphics[width=0.1\linewidth,valign=m]{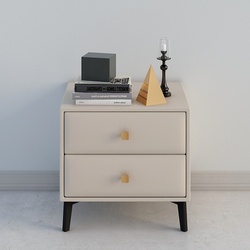} &
        \includegraphics[width=0.1\linewidth,valign=m]{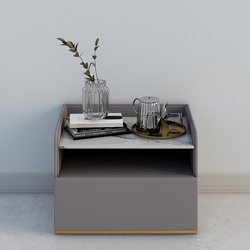} &
        \includegraphics[width=0.1\linewidth,valign=m]{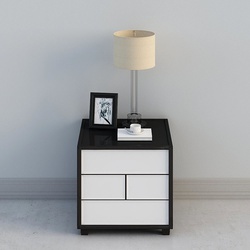} &
        \includegraphics[width=0.1\linewidth,valign=m]{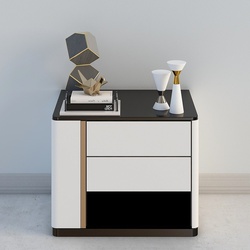} &
        \includegraphics[width=0.1\linewidth,valign=m]{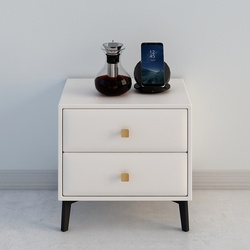} &
        \includegraphics[width=0.1\linewidth,valign=m]{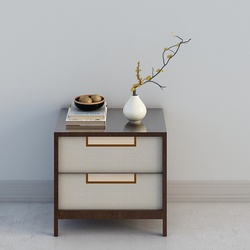} &
        \includegraphics[width=0.1\linewidth,valign=m]{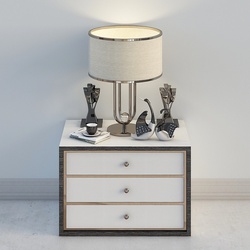} &
        \includegraphics[width=0.1\linewidth,valign=m]{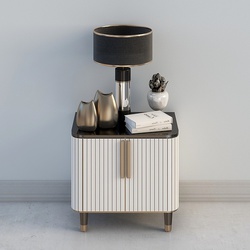} \tabularnewline
        wardrobe &
        \includegraphics[width=0.1\linewidth,valign=m]{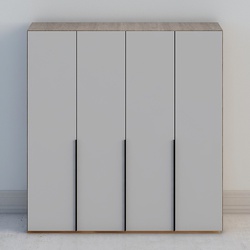} &
        \includegraphics[width=0.1\linewidth,valign=m]{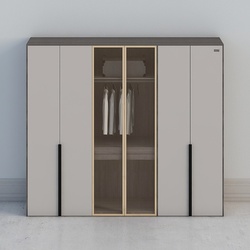} &
        \includegraphics[width=0.1\linewidth,valign=m]{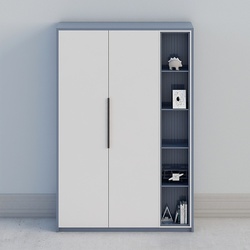} &
        \includegraphics[width=0.1\linewidth,valign=m]{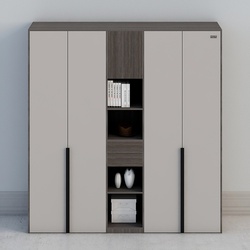} &
        \includegraphics[width=0.1\linewidth,valign=m]{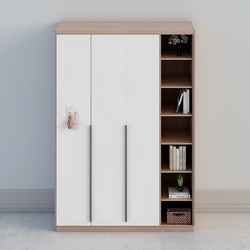} &
        \includegraphics[width=0.1\linewidth,valign=m]{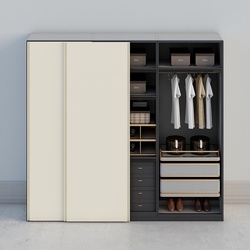} &
        \includegraphics[width=0.1\linewidth,valign=m]{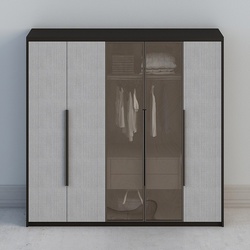} &
        \includegraphics[width=0.1\linewidth,valign=m]{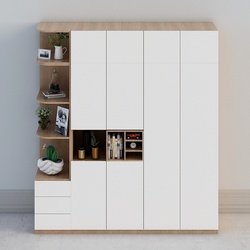} \tabularnewline
        chair &
        \includegraphics[width=0.1\linewidth,valign=m]{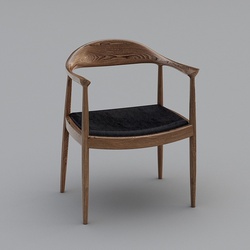} &
        \includegraphics[width=0.1\linewidth,valign=m]{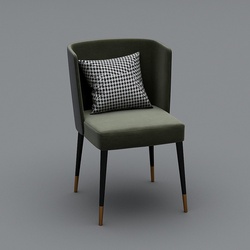} &
        \includegraphics[width=0.1\linewidth,valign=m]{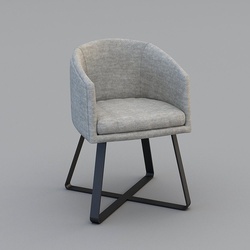} &
        \includegraphics[width=0.1\linewidth,valign=m]{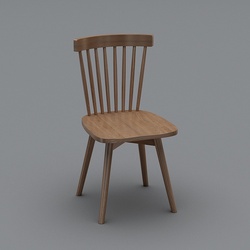} &
        \includegraphics[width=0.1\linewidth,valign=m]{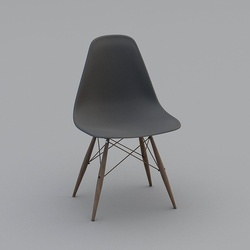} &
        \includegraphics[width=0.1\linewidth,valign=m]{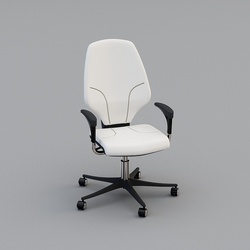} &
        \includegraphics[width=0.1\linewidth,valign=m]{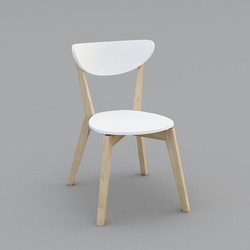} &
        \includegraphics[width=0.1\linewidth,valign=m]{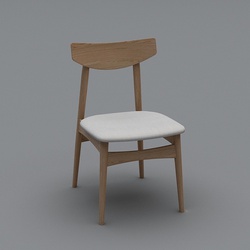} \tabularnewline
        table &
        \includegraphics[width=0.1\linewidth,valign=m]{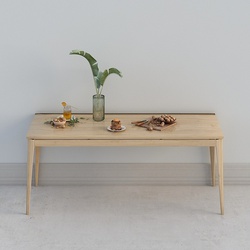} &
        \includegraphics[width=0.1\linewidth,valign=m]{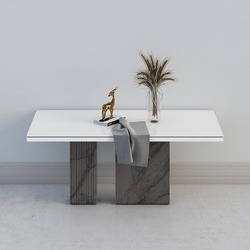} &
        \includegraphics[width=0.1\linewidth,valign=m]{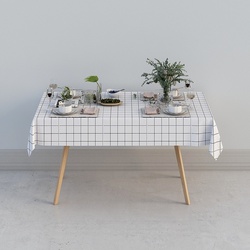} &
        \includegraphics[width=0.1\linewidth,valign=m]{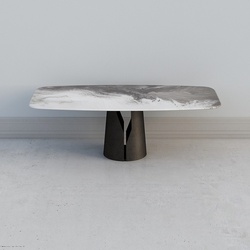} &
        \includegraphics[width=0.1\linewidth,valign=m]{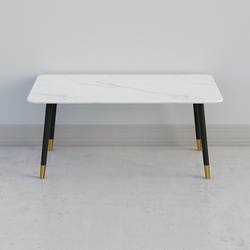} &
        \includegraphics[width=0.1\linewidth,valign=m]{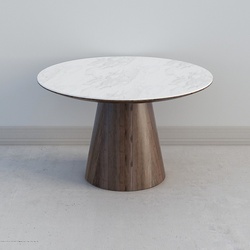} &
        \includegraphics[width=0.1\linewidth,valign=m]{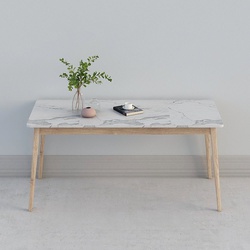} &
        \includegraphics[width=0.1\linewidth,valign=m]{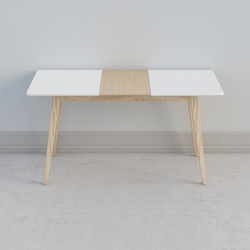} \tabularnewline
        desk &
        \includegraphics[width=0.1\linewidth,valign=m]{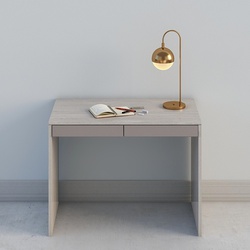} &
        \includegraphics[width=0.1\linewidth,valign=m]{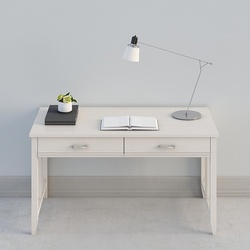} &
        \includegraphics[width=0.1\linewidth,valign=m]{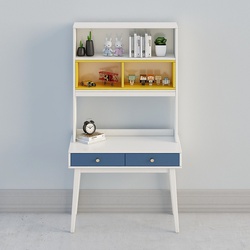} &
        \includegraphics[width=0.1\linewidth,valign=m]{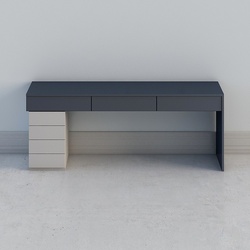} &
        \includegraphics[width=0.1\linewidth,valign=m]{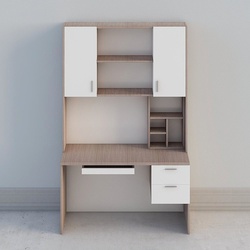} &
        \includegraphics[width=0.1\linewidth,valign=m]{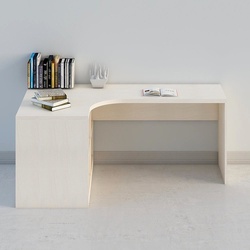} &
        \includegraphics[width=0.1\linewidth,valign=m]{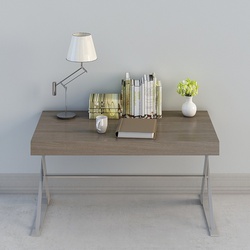} &
        \includegraphics[width=0.1\linewidth,valign=m]{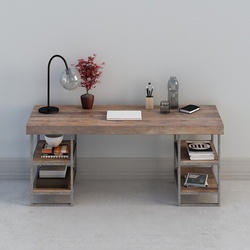} \tabularnewline
        shoe cabinet &
        \includegraphics[width=0.1\linewidth,valign=m]{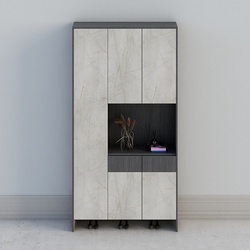} &
        \includegraphics[width=0.1\linewidth,valign=m]{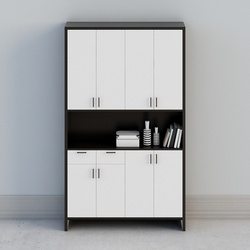} &
        \includegraphics[width=0.1\linewidth,valign=m]{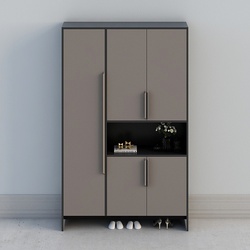} &
        \includegraphics[width=0.1\linewidth,valign=m]{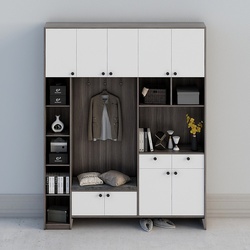} &
        \includegraphics[width=0.1\linewidth,valign=m]{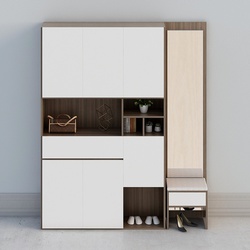} &
        \includegraphics[width=0.1\linewidth,valign=m]{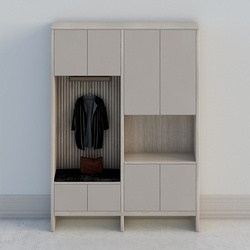} &
        \includegraphics[width=0.1\linewidth,valign=m]{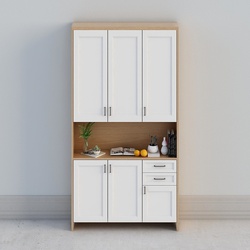} &
        \includegraphics[width=0.1\linewidth,valign=m]{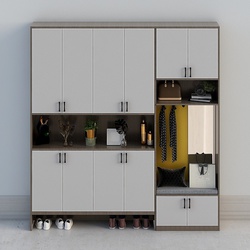} \tabularnewline
        chandelier &
        \includegraphics[width=0.1\linewidth,valign=m]{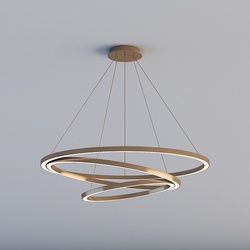} &
        \includegraphics[width=0.1\linewidth,valign=m]{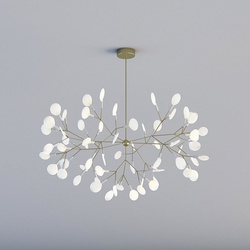} &
        \includegraphics[width=0.1\linewidth,valign=m]{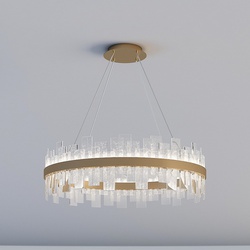} &
        \includegraphics[width=0.1\linewidth,valign=m]{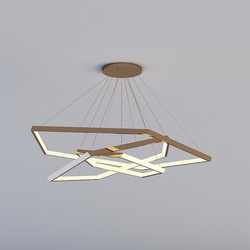} &
        \includegraphics[width=0.1\linewidth,valign=m]{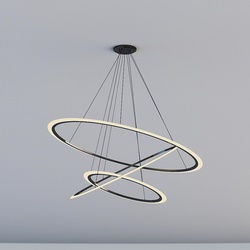} &
        \includegraphics[width=0.1\linewidth,valign=m]{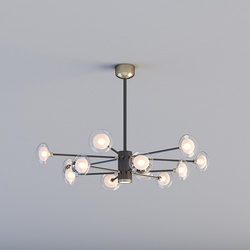} &
        \includegraphics[width=0.1\linewidth,valign=m]{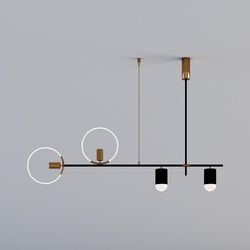} &
        \includegraphics[width=0.1\linewidth,valign=m]{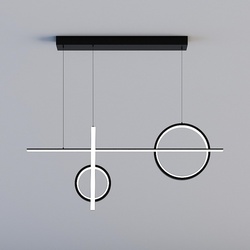} \tabularnewline  
        dressing table &
        \includegraphics[width=0.1\linewidth,valign=m]{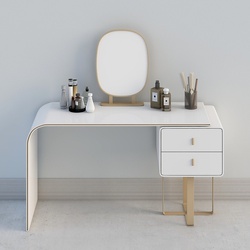} &
        \includegraphics[width=0.1\linewidth,valign=m]{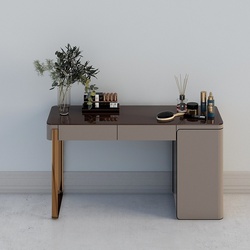} &
        \includegraphics[width=0.1\linewidth,valign=m]{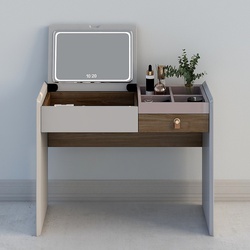} &
        \includegraphics[width=0.1\linewidth,valign=m]{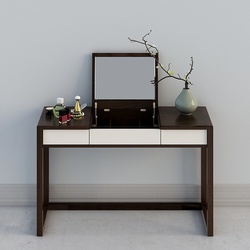} &
        \includegraphics[width=0.1\linewidth,valign=m]{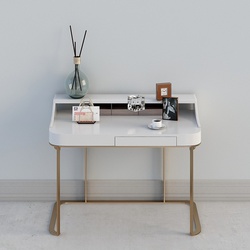} &
        \includegraphics[width=0.1\linewidth,valign=m]{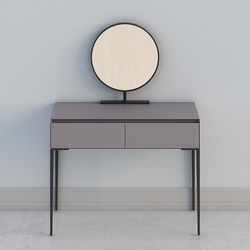} &
        \includegraphics[width=0.1\linewidth,valign=m]{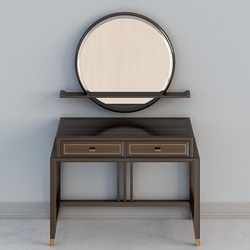} &
        \includegraphics[width=0.1\linewidth,valign=m]{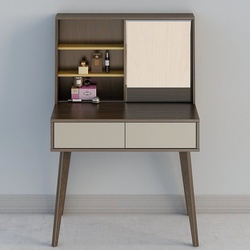} \tabularnewline        
    \end{tabular}
    \caption{Example objects in \method dataset. Note that each category has a diverse set of objects with high quality in both geometry and appearance.}
    \label{fig:visual-objects}
\end{figure}

{\bf Dataset visualizations.} We provide visualizations of the ground-truth point cloud and layout annotations in \cref{fig:visual-layout} to showcase the quality of the photo-realistic scans and the realism of the layout. Additionally, we show rendered images of objects in \cref{fig:visual-objects} to illustrate the diversity and quality of the objects in our dataset.

\section{Implementation Details of \method}
\label{sec:supp:imp}

\begin{figure}[t]
    \begin{tcolorbox}[width=\linewidth, colback=white, colframe=black, left=10pt, right=10pt, top=10pt, bottom=10pt]
        \begin{minipage}[t]{0.4\linewidth}
            {\color{prompt}\textbf{Prompt:}} \\
            \includegraphics[width=\linewidth]{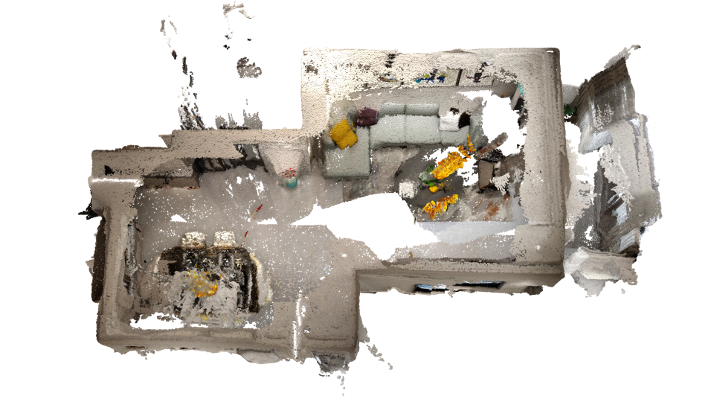} 
            {\color{img}\textbf{<point\_cloud>}} Detect walls, doors, windows, boxes. The reference code is as follows:
\begin{lstlisting}
@dataclass
class Wall:
    a_x: int
    a_y: int
    a_z: int
    b_x: int
    b_y: int
    b_z: int
    height: int
\end{lstlisting}
\begin{lstlisting}
@dataclass
class Door:
    wall_id: str
    position_x: int
    position_y: int
    position_z: int
    width: int
    height: int
\end{lstlisting}
\begin{lstlisting}
@dataclass
class Window:
    wall_id: str
    position_x: int
    position_y: int
    position_z: int
    width: int
    height: int
\end{lstlisting}
\begin{lstlisting}
@dataclass
class Bbox:
    <\color{black}{class}>: str
    position_x: int
    position_y: int
    position_z: int
    angle_z: int
    scale_x: int
    scale_y: int
    scale_z: int
\end{lstlisting}
        \end{minipage}
        \hfill
        \begin{minipage}[t]{0.56\linewidth}
            {\color{prompt}\textbf{Response:}} \\
\begin{lstlisting}
wall_0=Wall(144,15,30,139,15,30,101)
wall_1=Wall(144,15,30,144,26,30,101)
wall_2=Wall(139,15,30,128,16,30,101)
wall_3=Wall(128,16,30,118,18,30,101)
wall_4=Wall(47,18,30,118,18,30,101)
wall_5=Wall(47,18,30,47,42,30,101)
wall_6=Wall(144,26,30,152,26,30,101)
wall_7=Wall(152,26,30,152,42,30,101)
wall_8=Wall(47,42,30,67,42,30,101)
wall_9=Wall(67,42,30,67,53,30,101)
wall_10=Wall(138,42,30,152,42,30,101)
wall_11=Wall(138,42,30,138,53,30,101)
wall_12=Wall(49,53,30,67,53,30,101)
wall_13=Wall(49,53,30,49,151,30,101)
wall_14=Wall(138,53,30,152,53,30,101)
wall_15=Wall(152,53,30,152,170,30,101)
wall_16=Wall(19,151,30,49,151,30,101)
wall_17=Wall(19,151,30,19,269,30,101)
wall_18=Wall(118,170,30,152,170,30,101)
wall_19=Wall(118,170,30,118,269,30,101)
wall_20=Wall(19,269,30,118,269,30,101)
door_0=Door(wall_16,32,151,60,30,75)
door_1=Door(wall_17,19,194,60,27,75)
door_2=Door(wall_19,118,213,60,37,75)
door_3=Door(wall_20,98,269,60,30,75)
window_0=Window(wall_4,94,18,74,73,75)
window_1=Window(wall_17,19,162,60,17,75)
\end{lstlisting}
\begin{lstlisting}
bbox_0=Bbox(curtain,93,20,74,320,165,10,96)
bbox_1=Bbox(plants,144,46,53,320,27,27,74)
bbox_2=Bbox(air_conditioner,147,58,55,412,15,15,81)
bbox_3=Bbox(plants,58,61,52,480,24,31,70)
bbox_4=Bbox(tv_cabinet,55,103,41,480,111,19,34)
bbox_5=Bbox(tv,50,103,69,480,77,6,44)
bbox_6=Bbox(combination_sofa,119,110,45,480,166,105,48)
bbox_7=Bbox(painting,152,110,80,480,102,6,41)
bbox_8=Bbox(floor-standing_lamp,145,163,55,320,21,21,80)
bbox_9=Bbox(dining_chair,50,200,43,320,26,28,41)
bbox_10=Bbox(dining_chair,67,214,43,480,26,28,41)
bbox_11=Bbox(dining_table,49,228,50,480,77,40,43)
bbox_12=Bbox(dining_chair,67,233,43,480,26,28,41)
bbox_13=Bbox(painting,21,236,77,480,58,6,39)
bbox_14=Bbox(shoe_cabinet,113,246,68,480,52,16,120)
bbox_15=Bbox(dining_chair,41,252,43,480,26,28,41)
bbox_16=Bbox(dining_chair,67,252,43,480,26,28,41)
bbox_17=Bbox(sideboard,54,264,68,320,98,16,120)
\end{lstlisting}
        \end{minipage}
    \end{tcolorbox}
    \caption{An example conversation of \method.}
    \label{fig:converastion}
\end{figure}

In~\cref{fig:converastion}, we show the full prompt and an example response of \method. Note that we shift the point cloud and the corresponding layout and object box coordinates to ensure that they are all non-negative. Next, we quantize the coordinates into 1,280 bins, each with a resolution of $2.5\mathrm{cm}$. LLM predicts integer values, which are mapped back to continuous values using the inverse transformation and normalization, restoring them to the original coordinate system.

\begin{figure}[t]
    \centering
    \includegraphics[width=\linewidth]{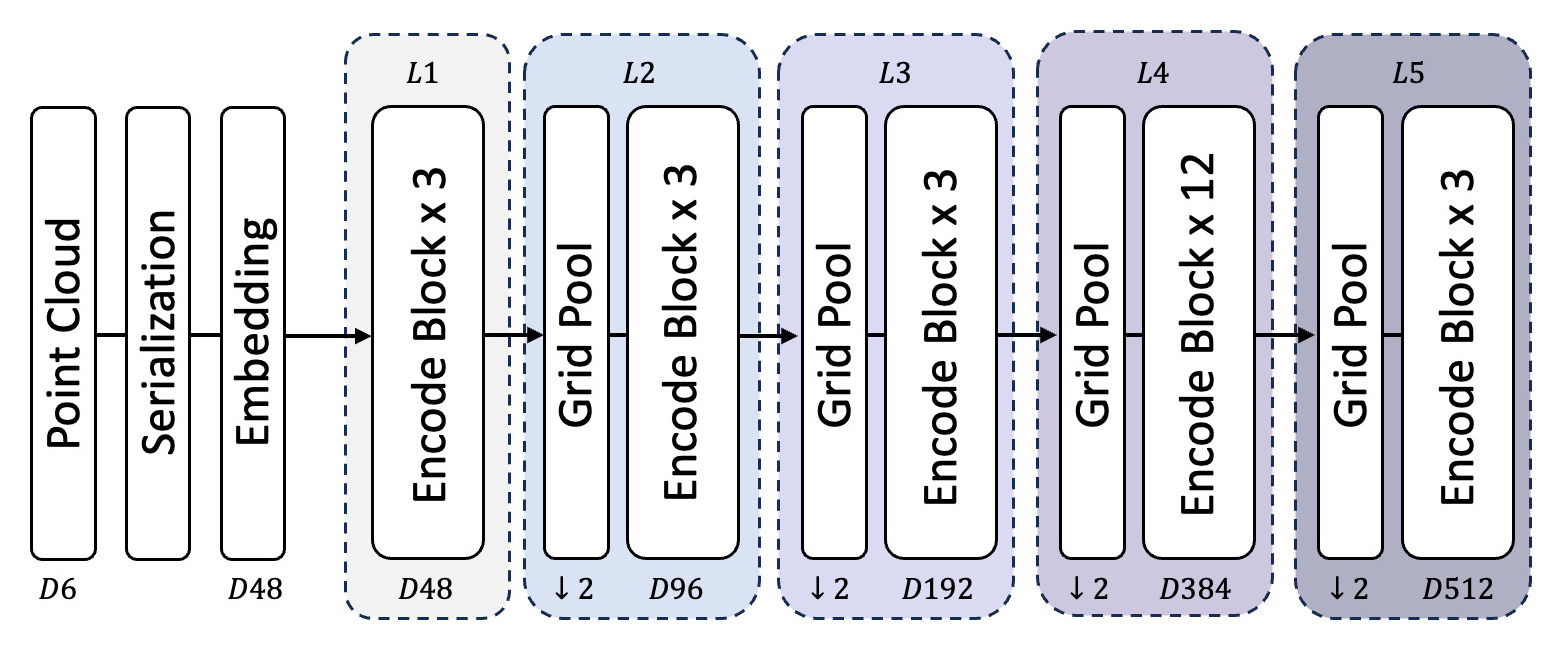}
    \caption{Five-level hierarchical structure of the point cloud encoder~\cite{Sonata}. $D$ indicates the feature dimension at each level, and grid pool downscales the point cloud resolution by 2 times along each dimension.}
    \label{fig:sonata-hierarchy}
\end{figure}

{\bf Data augmentations.} We apply a number of augmentations to the point cloud, which are summarized in \cref{tab:pcdaug} and explained below.

We begin with a cuboid crop following V-DETR~\cite{0001GYLLW00G24}. This process involves randomly selecting a point as the center of the cuboid and then determining random dimensions for the cuboid relative to the input point cloud size. We maintain a minimum aspect ratio of 0.8 and ensure that at least 50,000 points remain after the cropping process.

Then, we apply four types of jitter noises to the point cloud: (i) jitter that simulates the minor uncertainties inherent in sensor accuracy; (2) jitter that replicates noise in areas with greater uncertainty; (3) jitter that reflects noise along the edges, typically from floating points between foreground objects and the background; and (4) jitter that represents distant floating points outside the room, often attributed to glossy windows, reflective mirrors, and points along the path to the outdoors.

\begin{table}[ht]
    \centering
    \small
    \setlength{\tabcolsep}{2.5pt}
    \caption{Point cloud data augmentations.}
    \label{tab:pcdaug}
    \begin{tabular}{ll}
        \toprule
        Method & Parameters \tabularnewline
        \midrule
        cuboid crop & min\_points=50,000, aspect=0.8, min\_crop=0.75, max\_crop=1.0, $p$=0.1 \tabularnewline
        random jitter & $\sigma$=0.025, clip=0.05, ratio=0.8, $p$=0.9 \tabularnewline
        random jitter & $\sigma$=0.2, clip=0.2, ratio=0.05, $p$=0.8 \tabularnewline
        random jitter & $\sigma$=0.4, clip=1.0, ratio=0.001, $p$=0.75 \tabularnewline
        random jitter & $\sigma$=0.5, clip=4.0, ratio=0.0005, $p$=0.65 \tabularnewline
        elastic distort & params=[[0.2, 0.4], [0.8, 1.6]], $p$=[0.8, 0.5] \tabularnewline
        random rotate & axis=z, angle=[0, 90, 180, 270] \tabularnewline
        random scale & scale=[0.75, 1.25] \tabularnewline
        auto contrast & $p$=0.2 \tabularnewline
        chromatic translation & $p$=0.75, ratio=0.1 \tabularnewline
        chromatic jitter & std=0.05; $p$=0.8 \tabularnewline
        color drop & $p$=0.1 \tabularnewline
        \bottomrule
    \end{tabular}
\end{table}

{\bf Compute resources.} We train \method for 4 epochs with a total batch size of 64. The learning rate is set at $10^{-4}$, using a cosine scheduler with a warm-up ratio of 0.03. The parameters for AdamW optimizer are as follows: adam\_beta1 is 0.9, adam\_beta2 is 0.99, and adam\_epsilon is \(1 \times 10^{-8}\). We utilized 32 NVIDIA H20 GPUs, and training on \method dataset takes approximately one day.

For experiments on Structured3D~\cite{Structured3D} and ScanNet~\cite{ScanNet}, we use a batch size of 8. Fine-tuning on Structured3D involves 50 epochs, while keeping the other settings the same. For fine-tuning on ScanNet, we introduce an additional pre-training stage to mitigate the difference in label spaces across datasets. Specifically, we map the classes in our dataset to those of ScanNet, and further train the model on our dataset for 3 epochs, followed by fine-tuning on ScanNet for 30 epochs.

\lstset{%
basicstyle={\tiny\ttfamily},
numberstyle=\tiny,
showstringspaces=false,
language=Python,
keywordstyle=\color{blue},
mathescape=true,
escapeinside={<}{>},
aboveskip=0pt,
belowskip=0pt,
}

\section{Supplementary Results for \method Experiments}
\label{sec:supp:exp}

In this section, we present per-category quantitative results for (i) layout estimation, (ii) 3D object detection, and (iii) zero-shot detection on videos. 

\subsection{Layout Estimation}

{\bf Per-category results.} In \cref{tab:layout-supp}, we report the F1 scores in percentages at 0.25 and 0.5 thresholds of IoU$_\textrm{2D}$ for architectural elements \ie, walls, doors and windows, and the averaged score across the categories on Structured3D test set. 

\begin{table}[h]
    \centering
    \small
    \setlength{\tabcolsep}{3pt}
    \renewcommand{\arraystretch}{1.2}
    \caption{Experiment results on layout estimation.}
    \label{tab:layout-supp}
    \begin{tabular}{l|cccc|cccc}
        \toprule
        \multirow{2}{*}{Method}  & \multicolumn{4}{c|}{IoU$_\textrm{2D}$@0.25} & \multicolumn{4}{c}{IoU$_\textrm{2D}$@0.5} \tabularnewline
        \cmidrule(lr){2-5} \cmidrule(lr){6-9}
        & avg. & wall & door & window & avg. & wall & door & window \tabularnewline
        \midrule
        RoomFormer~\cite{YueKSE23}   & 83.4 & 86.0 & 85.1 & 78.9 & 81.4 & 84.6 & 83.1 & 76.6 \tabularnewline
        SceneScript~\cite{SceneScript}   & 90.4 & 92.9 & 92.8 & 85.5 & 89.2 & 91.9 & 92.5 & 83.3 \tabularnewline
        \midrule
        \method (ft.~Structured3D)  & 32.6 & 54.1 & 19.4 & 24.3 & 18.1 & 33.5 & 9.4 & 11.5 \tabularnewline
        \method (ft.~Ours) & 59.9 & 71.5 & 42.7 & 65.6 & 44.7 & 64.8 & 39.7 & 29.6 \tabularnewline
        \method (ft.~Ours $\rightarrow$ Structured3D) & \textbf{94.3} & \textbf{95.3} & \textbf{95.9} & \textbf{91.7} & \textbf{93.5} & \textbf{94.5} & \textbf{95.7} & \textbf{90.2} \tabularnewline
        \bottomrule
    \end{tabular}
\end{table}

\subsection{3D Object Detection}

{\bf Per-category results.} We report the F1 scores in percentages at 0.25 and 0.5 thresholds of IoU$_\textrm{3D}$ for 18 ScanNet categories and the averaged score across the categories in \cref{tab:object-25-supp} and \cref{tab:object-50-supp}, respectively.

For IoU$_\textrm{3D}$@0.25, we have +16.5\% overall gain compared to SceneScript, and +0.5\% difference compared to V-DETR. For IoU$_\textrm{3D}$@0.5, we have +15.8\% overall gain compared to SceneScript, and -4.2\% difference compared to V-DETR. The largest gaps between our model and V-DETR correspond to ``picture'', ``sink'', and ``shower\_curtain''. Note that these are either the smallest or the least occurring objects in ScanNet. 

\begin{table}[h]
    \centering
    \scriptsize
    \setlength{\tabcolsep}{2pt}
    \renewcommand{\arraystretch}{1.2}
    \caption{3D object detection results with IoU$_\textrm{3D}$@0.25. $^\dagger$: trained on ScanNet only. $^*$: trained on our dataset only. ``n/a'' indicates the category is not included in our dataset. }
    \label{tab:object-25-supp}
    \begin{tabular}{l|c|cccccccccccccccccc}
        \toprule
        Method & \rotatebox{90}{average} & \rotatebox{90}{bathtub} & \rotatebox{90}{bed} & \rotatebox{90}{bookshelf} & \rotatebox{90}{cabinet} & \rotatebox{90}{chair} & \rotatebox{90}{counter} & \rotatebox{90}{curtain} & \rotatebox{90}{desk} & \rotatebox{90}{door} & \rotatebox{90}{garbbin.} & \rotatebox{90}{picture} & \rotatebox{90}{refrige.} & \rotatebox{90}{s.curtain} & \rotatebox{90}{sink} & \rotatebox{90}{sofa} & \rotatebox{90}{table} & \rotatebox{90}{toilet} & \rotatebox{90}{window} \tabularnewline
        \midrule
        V-DETR~\cite{0001GYLLW00G24} & 65.1 & 77.4 & \textbf{80.1} & 48.8 & \textbf{44.4} & 82.5 & \textbf{55.6} & 66.1 & \textbf{68.0} & 62.3 & 54.0 & \textbf{47.4} & 50.8 & 74.0 & \textbf{79.5} & 72.7 & 55.3 & \textbf{98.0} & 54.9 \tabularnewline
        SceneScript~\cite{SceneScript} & 49.1 & 64.5 & 71.1 & 39.7 & 30.3 & 81.1 & 43.4 & 30.9 & 53.4 & 41.7 & 42.6 & 11.8 & 28.2 & 58.6 & 48.9 & 68.3 & 55.8 & 77.5 & 36.5 \tabularnewline
        \midrule
        \textsc{SpatialLM}$^\dagger$  & 2.9 & 6.5 & 7.0 & 2.5 & 1.2 & 4.0 & 1.4 & 2.3 & 4.8 & 2.5 & 0.3 & 0.0 & 1.4 & 2.3 & 0.9 & 1.7 & 4.7 & 5.0 & 3.7 \tabularnewline
        \textsc{SpatialLM}$^*$ & 33.8 & 57.0 & 49.2 & 24.4 & 12.2 & 50.9 & n/a & 15.3 & 48.1 & n/a & n/a & 9.7 & 23.4 & n/a & 3.6 & 38.3 & 31.4 & 75.6 & n/a \tabularnewline
        \method & \textbf{65.6} & \textbf{80.6} & 79.9 & \textbf{52.0} & 40.0 & \textbf{86.6} & 51.0 & \textbf{66.8} & 62.8 & \textbf{67.1} & \textbf{55.6} & 29.7 & \textbf{53.6} & \textbf{81.6} & 70.7 & \textbf{78.9} & \textbf{63.9} & 95.4 & \textbf{64.3} \tabularnewline 
        \bottomrule
    \end{tabular}
\end{table}

\begin{table}[h]
    \centering
    \scriptsize
    \setlength{\tabcolsep}{2pt}
    \renewcommand{\arraystretch}{1.2}
    \caption{3D object detection results with IoU$_\textrm{3D}$@0.5. $^\dagger$: trained on ScanNet only. $^*$: trained on our dataset only. ``n/a'' indicates the category is not included in our dataset.}
    \label{tab:object-50-supp}
    \begin{tabular}{l|c|cccccccccccccccccc}
        \toprule
        Method & \rotatebox{90}{average} & \rotatebox{90}{bathtub} & \rotatebox{90}{bed} & \rotatebox{90}{bookshelf} & \rotatebox{90}{cabinet} & \rotatebox{90}{chair} & \rotatebox{90}{counter} & \rotatebox{90}{curtain} & \rotatebox{90}{desk} & \rotatebox{90}{door} & \rotatebox{90}{garbbin.} & \rotatebox{90}{picture} & \rotatebox{90}{refrige.} & \rotatebox{90}{s.curtain} & \rotatebox{90}{sink} & \rotatebox{90}{sofa} & \rotatebox{90}{table} & \rotatebox{90}{toilet} & \rotatebox{90}{window} \tabularnewline
        \midrule
        V-DETR~\cite{0001GYLLW00G24} & \textbf{56.8} & 68.8 & \textbf{75.5} & \textbf{50.4} & \textbf{39.1} & 76.2 & \textbf{43.6} & \textbf{54.1} & 52.1 & \textbf{50.9} & \textbf{47.8} & \textbf{42.7} & \textbf{49.2} & \textbf{61.5} & \textbf{61.4} & 65.4 & 50.6 & 88.6 & 43.9 \tabularnewline
        SceneScript~\cite{SceneScript} & 36.8 & 54.8 & 65.9 & 37.3 & 21.2 & 73.7 & 25.3 & 16.9 & 44.5 & 23.5 & 30.5 & 3.0 & 23.3 & 19.5 & 24.9 & 62.0 & 50.1 & 64.2 & 21.7 \tabularnewline
        \midrule
        \textsc{SpatialLM}$^\dagger$ & 0.7 & 2.2 & 2.0 & 0.0 & 0.1 & 0.6 & 1.4 & 0.8 & 1.7 & 0.4 & 0.0 & 0.0 & 1.4 & 0.0 & 0.0 & 0.0 & 0.2 & 2.0 & 0.0 \tabularnewline
        \textsc{SpatialLM}$^*$ & 22.6 & 37.6 & 41.0 & 13.5 & 6.8 & 30.7 & n/a & 1.0 & 34.3 & n/a & n/a & 2.6 & 17.7 & n/a & 1.1 & 28.4 & 19.5 & 59.0 & n/a \tabularnewline
        \method & 52.6 & \textbf{74.2} & 71.7 & 47.7 & 29.5 & \textbf{79.3} & 40.1 & 53.3 & \textbf{52.6} & 46.6 & 41.7 & 10.2 & 42.9 & 50.6 & 40.2 & \textbf{71.1} & \textbf{60.3} & \textbf{90.5} & \textbf{45.0} \tabularnewline
        \bottomrule
    \end{tabular}
\end{table}

\subsection{Zero-shot Detection on Videos}

{\bf Per-category results.} We collect 107 virtual room-tour videos and use MASt3R-SLAM~\cite{MAST3RSLAM} to reconstruct a point cloud from each video. Note that these videos are rendered from virtual scenes created by professional designers, thus ground-truth 3D annotations are available. To calibrate the reconstructed point cloud with ground truth 3D annotations, we align the estimated camera trajectory from MASt3R-SLAM to the ground-truth camera trajectory. Then, we perform zero-shot detection with \method on the calibrated point clouds. 

In \cref{tab:zeroshot}, we report the quantitative results on the top-20 occurring categories in the video test set. Note that this is a challenging task, because the reconstructed point clouds are often noisy and highly incomplete. Furthermore, because of the differences in scale and position due to imperfect alignment of camera trajectories, the objects in the point cloud may not match well with the ground truth 3D annotations, leading to lower F1 scores.

\begin{table}[h]
    \renewcommand{\arraystretch}{1.2}
    \caption{Zero-shot results on layout estimation and 3D object detection with IoU@0.25.}
    \label{tab:zeroshot}
    \begin{subtable}[t]{1.0\textwidth}
        \small
        \centering
        \setlength{\tabcolsep}{5pt}
        \caption{Layout estimation}
        \label{subtab:layout}
        \begin{tabular}{c|ccc}
            \toprule
            avg. & wall & door & window \tabularnewline
            \midrule
            55.7 & 68.2 & 47.4 & 51.4 \tabularnewline
            \bottomrule
        \end{tabular}
    \end{subtable}
    \hfill
    \begin{subtable}[t]{1.0\textwidth}
        \centering
        \small
        \setlength{\tabcolsep}{1.6pt}
        \caption{3D object detection}
        \label{subtab:detection}
        \begin{tabular}{c|cccccccccccccccccccc}
            \toprule
            \rotatebox{90}{average} & \rotatebox{90}{curtain} & \rotatebox{90}{nightstand} & \rotatebox{90}{chandelier} & \rotatebox{90}{wardrobe} & \rotatebox{90}{bed} & \rotatebox{90}{sofa} & \rotatebox{90}{chair} & \rotatebox{90}{cabinet} & \rotatebox{90}{diningtab.} & \rotatebox{90}{plants} & \rotatebox{90}{tvcab.} & \rotatebox{90}{coffeetab.} & \rotatebox{90}{sidetab.} & \rotatebox{90}{aircond.} & \rotatebox{90}{dresser} & \rotatebox{90}{stool} & \rotatebox{90}{refrige.} & \rotatebox{90}{painting} & \rotatebox{90}{carpet} & \rotatebox{90}{tv} \tabularnewline
            \midrule
            36.9 & 37.0 & 67.0 & 36.8 & 39.6 & 95.2 & 69.1 & 32.3 & 11.2 & 24.2 & 26.3 & 27.3 & 64.9 & 9.7 & 24.0 & 46.7 & 30.8 & 16.7 & 38.2 & 24.1 & 18.0 \tabularnewline
            \bottomrule
        \end{tabular}
    \end{subtable}
\end{table}

\section{Extensions of \method}
\label{sec:supp:ext}

In this section, we present more proof-of-concept experiment results on future extensions of \method, including (i) handling point clouds from diverse sources, and (ii) adapting to various downstream tasks via language-based prompts.

\subsection{Supporting Point Clouds from Diverse Sources}

Point clouds offer a lightweight and flexible 3D representation that can be readily generated from a variety of sources. Meanwhile, our dataset is large and diverse, enabling strong zero-shot performance over point cloud inputs in both synthetic and real-world data. In this experiment, we examine the following four types of input sources:

\begin{itemize}
    \item {\bf Text-to-3D}: We first generate an image via Flux-dev~\cite{flux2024} with prompt ``a low poly 3D living room in cartoon style'', then convert the image into 3D using TRELLIS~\cite{trellis}. 
    
    \item {\bf Hand-held video camera}: We take a real-world video using a hand-held camera and perform 3D reconstruction using MASt3R-SLAM~\cite{MAST3RSLAM}.
    
    \item {\bf LiDAR-based reconstruction}: We take point clouds reconstructed from captures using iPhone ARKit from MultiScan~\cite{MaoZJCS22}.
    
    \item {\bf Synthetic mesh}: We create point clouds by randomly sampling mesh surfaces from a synthetic dataset HSSD~\cite{HSSD}.
\end{itemize}

\begin{figure}[h]
    \centering
    \setlength{\tabcolsep}{2pt}
    \begin{tabular}{c|ccc}
        \includegraphics[width=0.25\linewidth,valign=m]{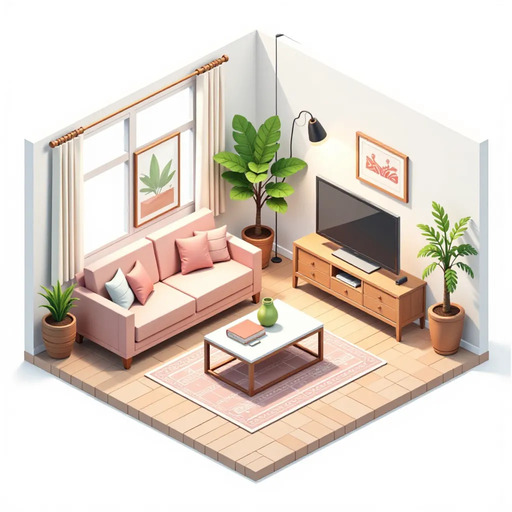} &
        \includegraphics[width=0.3\linewidth,valign=m]{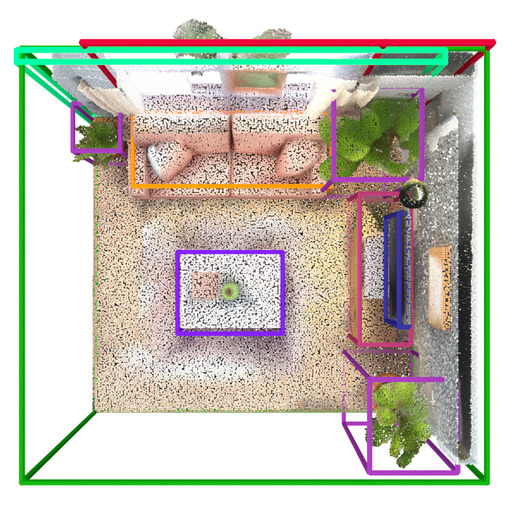} &
        \includegraphics[width=0.2\linewidth,valign=m]{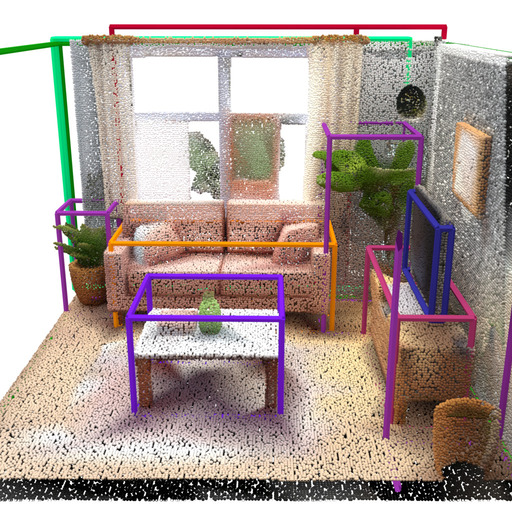} &
        \includegraphics[width=0.2\linewidth,valign=m]{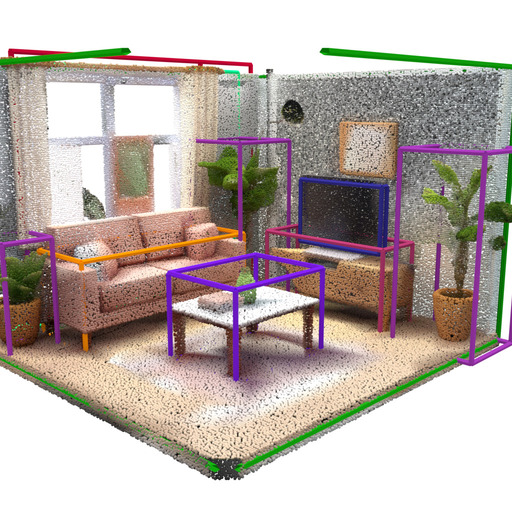} \tabularnewline
        \includegraphics[width=0.25\linewidth,valign=m]{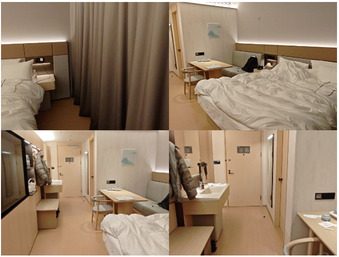} &
        \includegraphics[width=0.3\linewidth,valign=m]{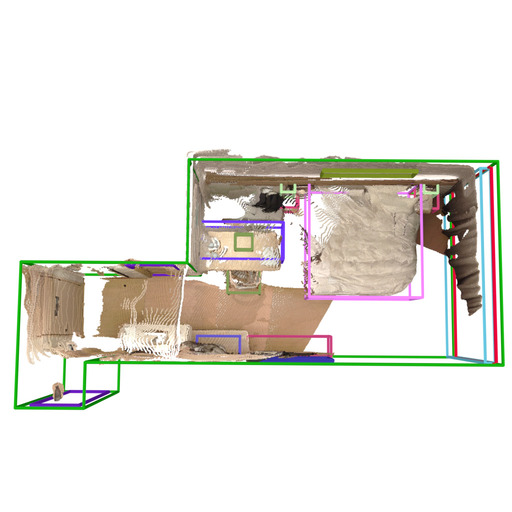} &
        \includegraphics[width=0.2\linewidth,valign=m]{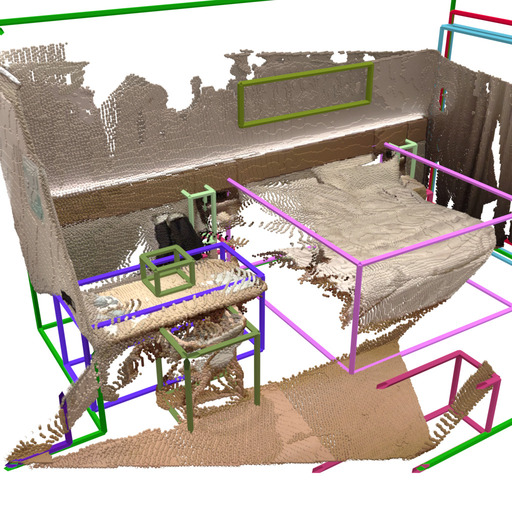} &
        \includegraphics[width=0.2\linewidth,valign=m]{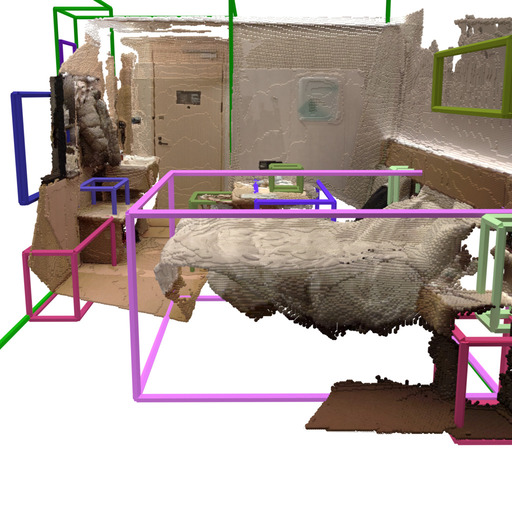} \tabularnewline
        \includegraphics[width=0.25\linewidth,valign=m]{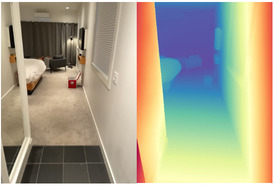} &
        \includegraphics[width=0.3\linewidth,valign=m]{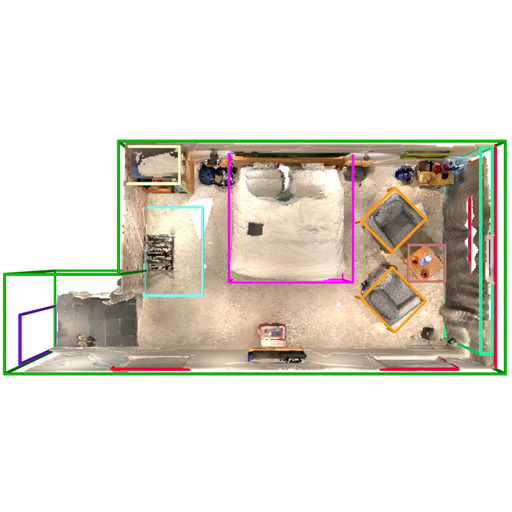} &
        \includegraphics[width=0.2\linewidth,valign=m]{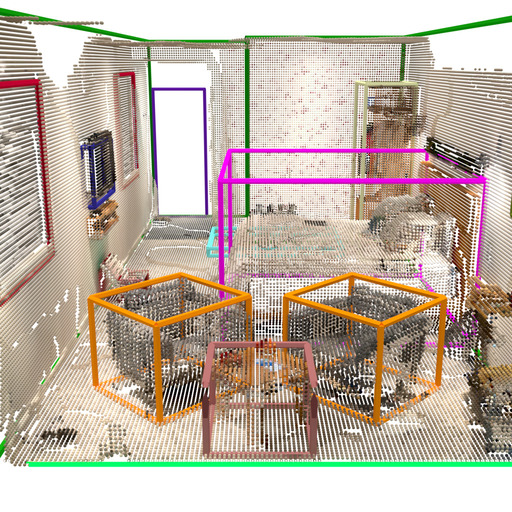} &
        \includegraphics[width=0.2\linewidth,valign=m]{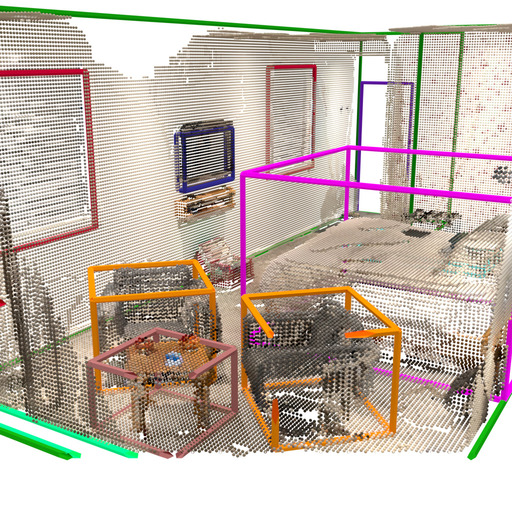} \tabularnewline
        \includegraphics[width=0.25\linewidth,valign=m]{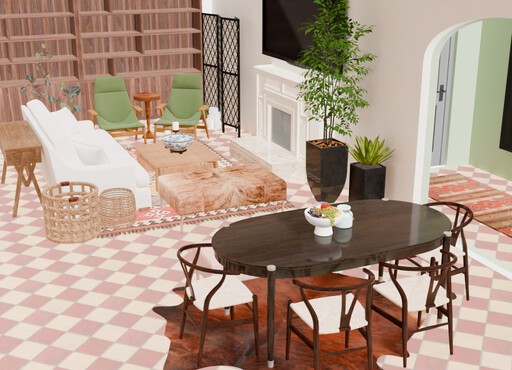} &
        \includegraphics[width=0.3\linewidth,valign=m]{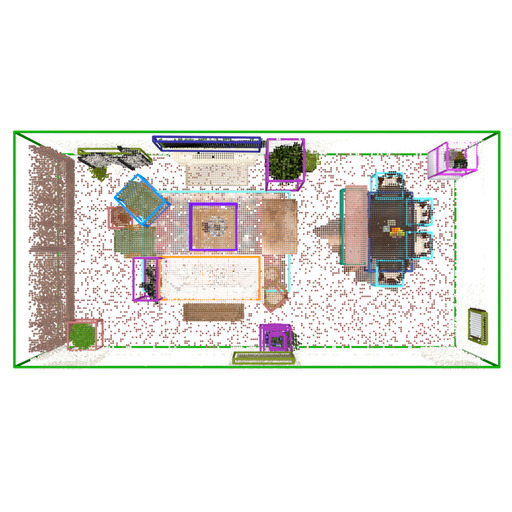} &
        \includegraphics[width=0.2\linewidth,valign=m]{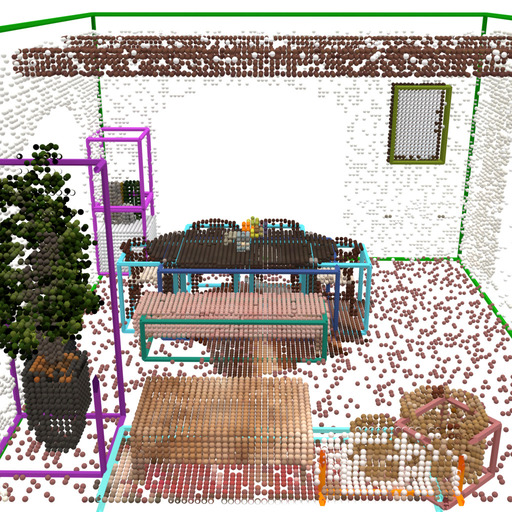} &
        \includegraphics[width=0.2\linewidth,valign=m]{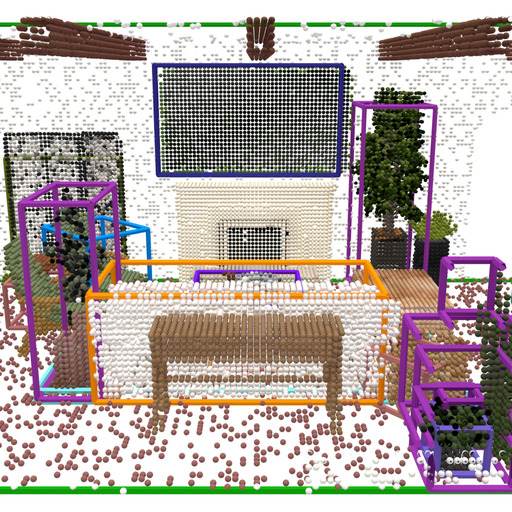} \tabularnewline
    \end{tabular}
    \caption{Qualitative results on various input sources. {\bf First row}: Text-to-3D. {\bf Second row}: Hand-held video camera. {\bf Third row}: LiDAR-based reconstruction. {\bf Fourth row}: Synthetic mesh.}
    \label{fig:zeroshot-supp}
\end{figure}

As can be seen in \cref{fig:zeroshot-supp}, our model is robust to variations in point cloud origin, structure, and appearance.

\subsection{Task Adaptation via Language-based Prompts}

A key idea of our work is to represent the 3D scene structures in pure text form. This enables \method to be conveniently adapted to different downstream tasks via natural language instruction, without changing the underlying model architecture. Below, we demonstrate two such extensions.

\begin{figure}[h]
    \centering
    \setlength{\tabcolsep}{2pt}
    \begin{tabular}{cccc}
        \includegraphics[width=0.24\linewidth,valign=m]{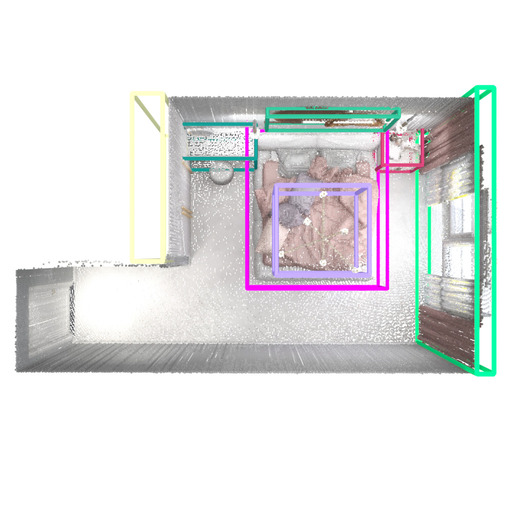} &
        \includegraphics[width=0.24\linewidth,valign=m]{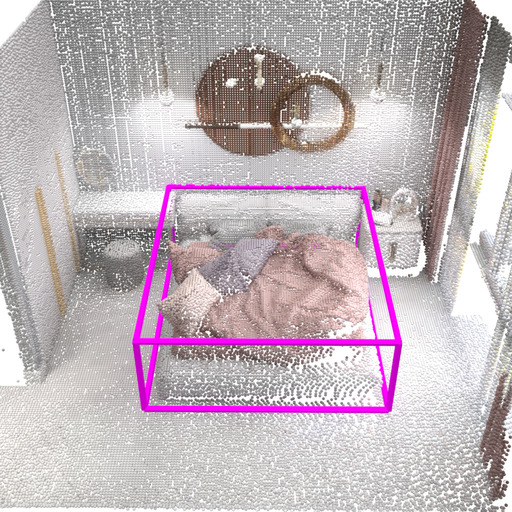} &
        \includegraphics[width=0.24\linewidth,valign=m]{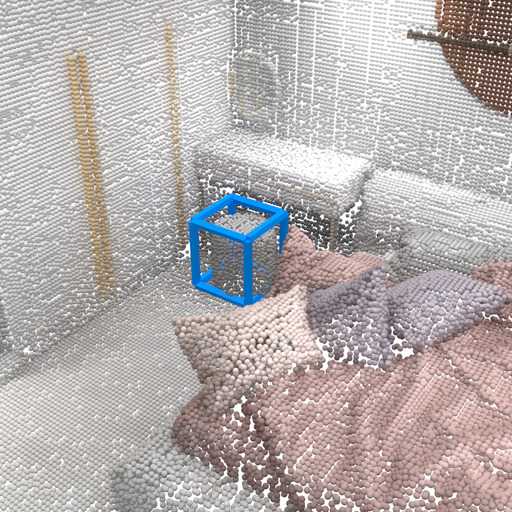} &
        \includegraphics[width=0.24\linewidth,valign=m]{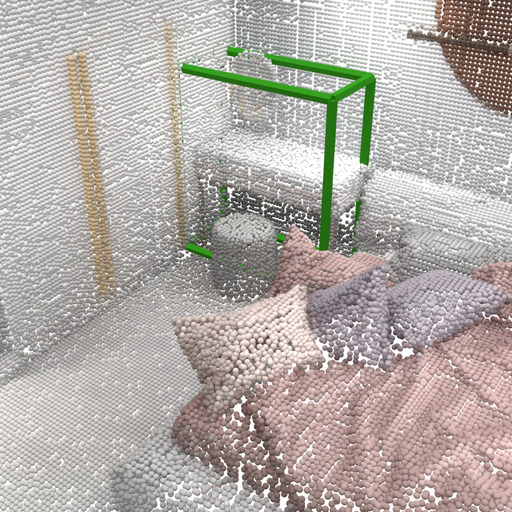} \tabularnewline
        all objects & bed & chair & desk \tabularnewline
        \includegraphics[width=0.24\linewidth,valign=m]{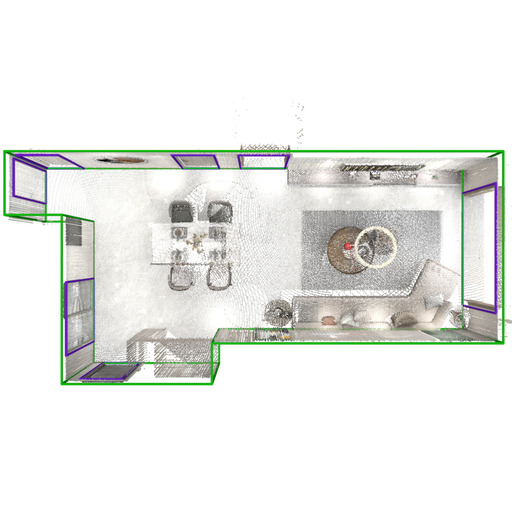} &
        \includegraphics[width=0.24\linewidth,valign=m]{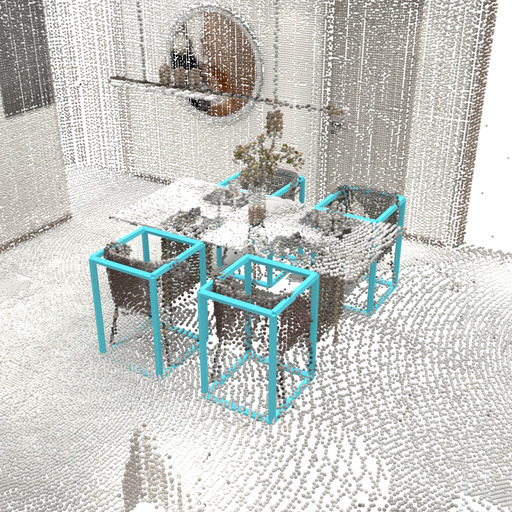} &
        \includegraphics[width=0.24\linewidth,valign=m]{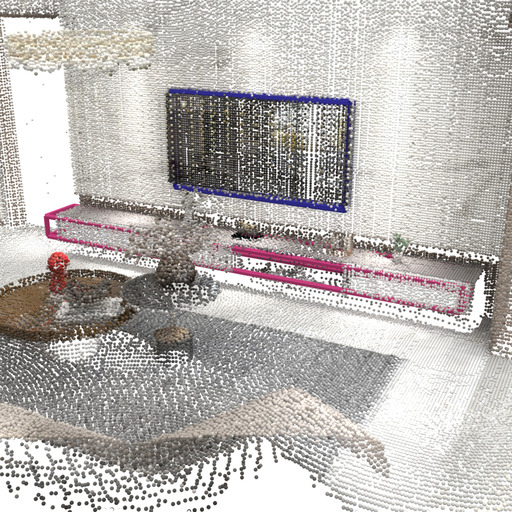} &
        \includegraphics[width=0.24\linewidth,valign=m]{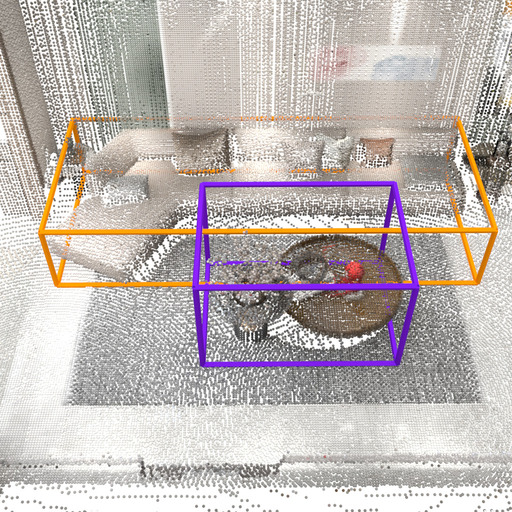} \tabularnewline
        layout & dining\_chair & tv, tv\_cabinet & sofa, coffee\_table \tabularnewline
    \end{tabular}
    \caption{Qualitative results of detection with user-specified categories. Labels below each figure indicate the category specified for detection.}
    \label{fig:prompt-supp}
\end{figure}

{\bf Detection with user-specified categories.} We enable the model to perform object detection conditioned on user-specified categories by leveraging the flexibility of LLMs. During training, the input prompt is customized to indicate which object categories to detect. The model then learns to selectively predict oriented bounding boxes (OBBs) for instances matching the specified categories, all within a unified training procedure. This experiment highlights the model's adaptability through prompt-based control. Some qualitative results are provided in \cref{fig:prompt-supp}.

{\bf Semantic label completion.} Besides natural language prompts, \method also supports structured language scripts as input, enabling novel tasks such as semantic label completion. In this task, the model receives object bounding boxes and their poses written in a general-purpose language (\ie, Python), while the object categories are left unspecified. The model then predicts the semantic label of each object based on the spatial layout and the corresponding point cloud data. Note that this problem often arises in real-world design workflows that involve 3D assets with known positions but without semantic labels.

\begin{figure}[h]
    \centering
    \setlength{\tabcolsep}{2pt}
    \begin{tabular}{ccc}
        \includegraphics[width=0.3\linewidth,valign=m]{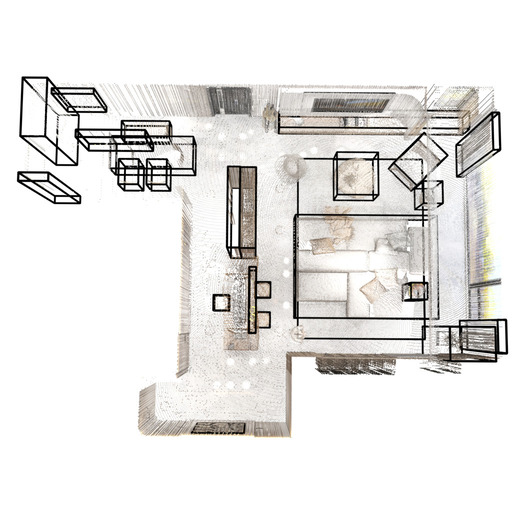} &
        \includegraphics[width=0.3\linewidth,valign=m]{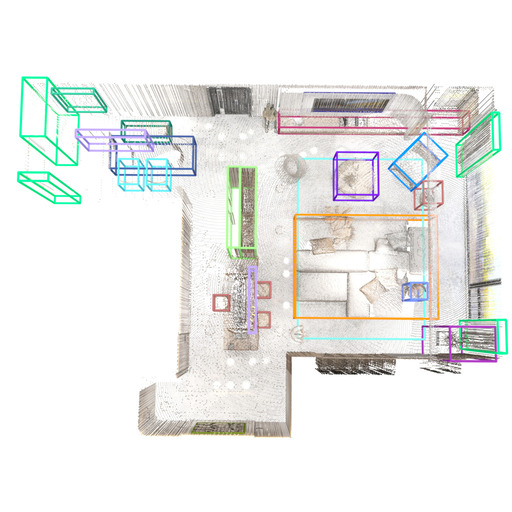} &
        \includegraphics[width=0.3\linewidth,valign=m]{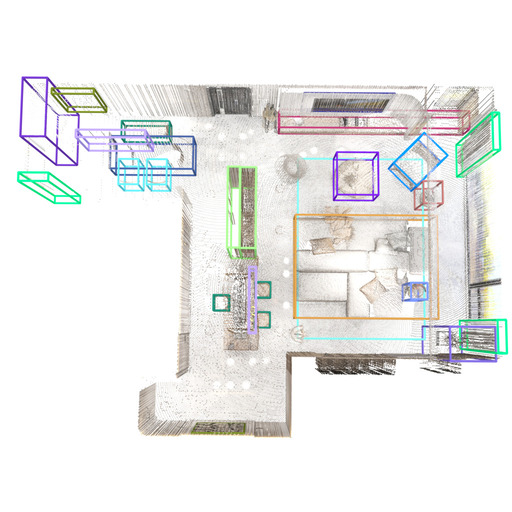} \tabularnewline
        \includegraphics[width=0.3\linewidth,valign=m]{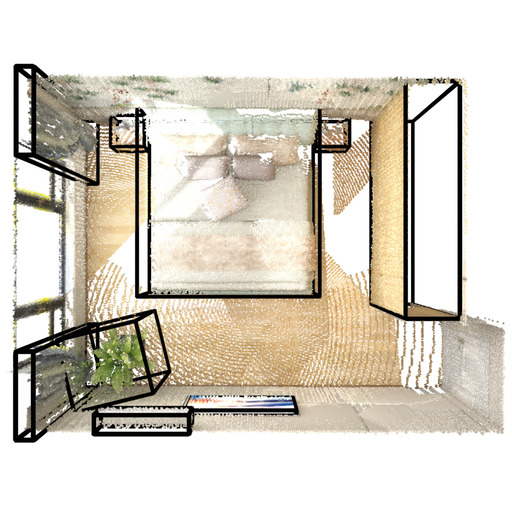} &
        \includegraphics[width=0.3\linewidth,valign=m]{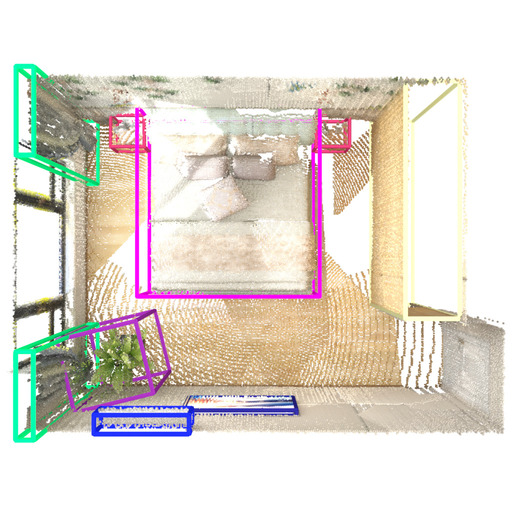} &
        \includegraphics[width=0.3\linewidth,valign=m]{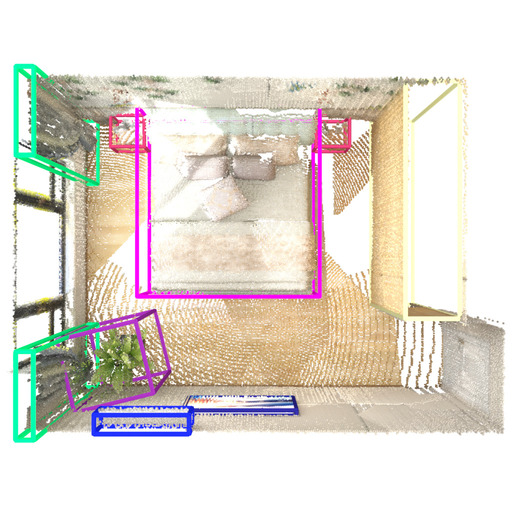} \tabularnewline
        \includegraphics[width=0.3\linewidth,valign=m]{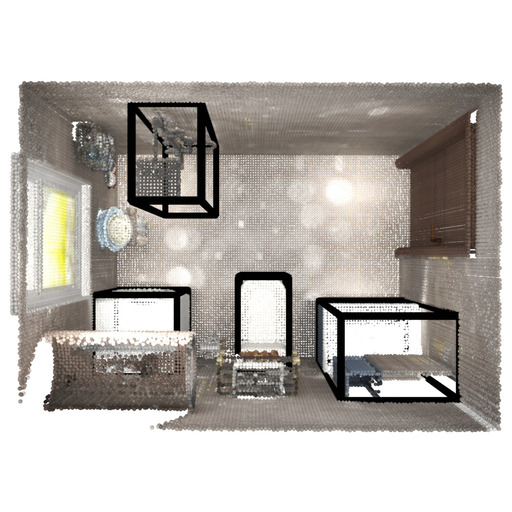} &
        \includegraphics[width=0.3\linewidth,valign=m]{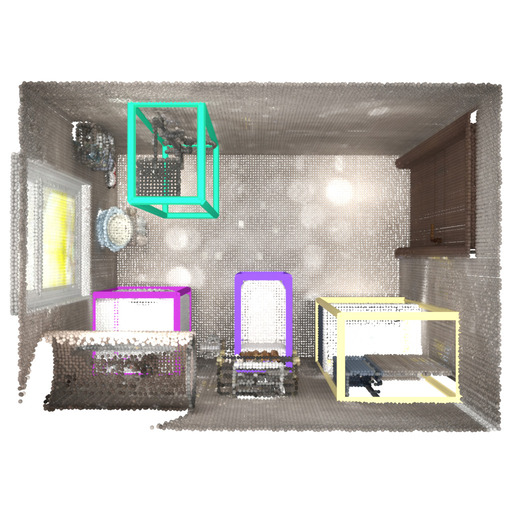} &
        \includegraphics[width=0.3\linewidth,valign=m]{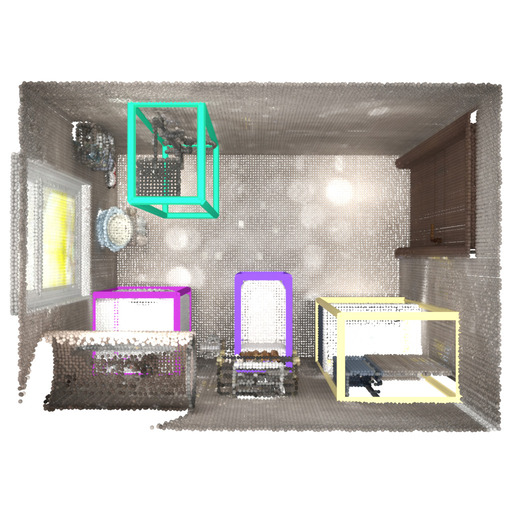} \tabularnewline
        Input & GT & \method \tabularnewline
    \end{tabular}
    \caption{Qualitative results of semantic label completion. Black-colored input boxes denote objects with unknown categories, which are classified by \method.}
    \label{fig:complete-supp}
\end{figure}

Compared to 3D object detection, this task is relatively easy. Nevertheless, it demonstrates the versatility and potential use of large language models for structured 3D understanding. We show some qualitative results in \cref{fig:complete-supp}. Our model achieves an overall classification accuracy of 96.8\% on the test set of our dataset. 

\end{document}